\documentclass[10pt,twocolumn,letterpaper]{article}

\usepackage{iccv}
\usepackage{times}
\usepackage{epsfig}
\usepackage{graphicx}
\usepackage{amsmath}
\usepackage{amssymb}
\usepackage{graphicx}
\usepackage{booktabs}
\usepackage{color,xcolor}
\usepackage{colortbl}
\usepackage[accsupp]{axessibility} 

\usepackage[breaklinks=true,colorlinks,bookmarks=false]{hyperref}

\iccvfinalcopy 

\def\iccvPaperID{****} 

\ificcvfinal\fi

\begin{document}

\title{Self-supervised Cross-view Representation Reconstruction for \\ Change Captioning}

\author{Yunbin Tu\textsuperscript{1}, Liang Li\textsuperscript{2,6}\footnotemark[1], Li Su\textsuperscript{1,3}\footnotemark[1], Zheng-Jun Zha\textsuperscript{4}, Chenggang Yan\textsuperscript{5,6}, Qingming Huang\textsuperscript{1,2,3}  \\
\textsuperscript{1}University of Chinese Academy of Sciences, Beijing, China\\
\textsuperscript{2}Key Lab of Intelligent Information Processing, ICT, CAS, Beijing, China\\
\textsuperscript{3}Peng Cheng Laboratory, Shenzhen, China\\
\textsuperscript{4}University of Science and Technology of China, Hefei, China\\
\textsuperscript{5}Hangzhou Dianzi University, Hangzhou, China\\
\textsuperscript{6}Lishui Institute of Hangzhou Dianzi University, Hangzhou, China\\
{\tt\small tuyunbin22@mails.ucas.ac.cn, liang.li@ict.ac.cn, \{suli,qmhuang\}@ucas.ac.cn}
}


\maketitle
\renewcommand{\thefootnote}{\fnsymbol{footnote}}
\footnotetext[1]{Corresponding authors}
\ificcvfinal\thispagestyle{empty}\fi

\begin{abstract}
   Change captioning aims to describe the difference between a pair of similar images. Its key challenge is how to learn a stable difference representation under pseudo changes caused by viewpoint change. In this paper, we address this by proposing a self-supervised cross-view representation reconstruction (SCORER) network. Concretely, we first design a multi-head token-wise matching to model relationships between cross-view features from similar/dissimilar images. Then, by maximizing cross-view contrastive alignment of two similar images, SCORER learns two view-invariant image representations in a self-supervised way. Based on these, we reconstruct the representations of unchanged objects by cross-attention, thus learning a stable difference representation for caption generation. Further, we devise a cross-modal backward reasoning to improve the quality of caption. This module reversely models a ``hallucination'' representation with the caption and ``before'' representation. By pushing it closer  to the ``after'' representation, we enforce the caption to be informative about the difference  in a self-supervised manner.  Extensive experiments show our method achieves the state-of-the-art  results on four  datasets. The code is available at \url{https://github.com/tuyunbin/SCORER}.
\end{abstract}

\section{Introduction}

Change captioning is a new task of vision and language, which requires not only understanding the contents of two similar images, but also describing their difference with natural language. In real world, this task brings a variety of applications, such as generating elaborated reports about monitored facilities \cite{hoxha2022change,jhamtani2018learning} and pathological changes \cite{liu2021contrastive,li2023dynamic}. 

\begin{figure}[t]
  \centering
   \includegraphics[width=1\linewidth]{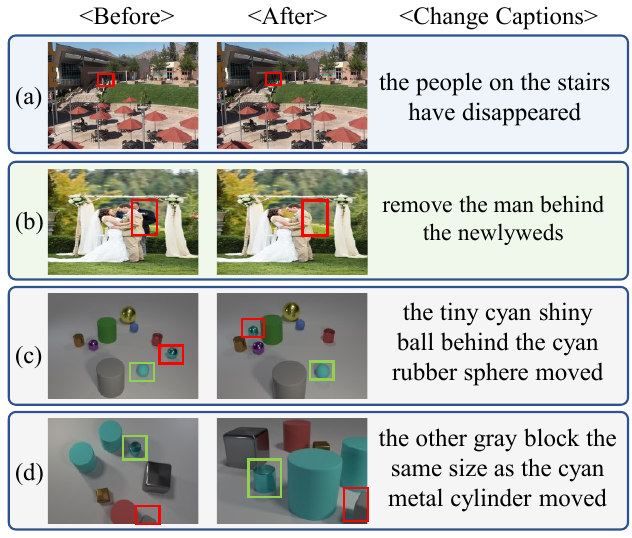}

   \caption{The examples of change captioning. (a) is from a surveillance scene with underlying illumination change. (b) is from an image editing scene. (c) shows that with both object move and moderate viewpoint change. (d) shows that with both object move and extreme viewpoint change. Changed objects and referents are shown in  red and green boxes, respectively.}
   \label{fig1}
\end{figure}

While single-image captioning is already regarded as a very challenging task, change captioning  carries additional difficulties. Simply locating inconspicuous differences is one such challenge (Fig. \ref{fig1} (a) (b)). Further, in a dynamic environment, it is common to acquire two images under different viewpoints, which leads to pseudo changes about objects' scale and location  (Fig. \ref{fig1} (c) (d)).  As such, change captioning needs to characterize the real change while resisting pseudo changes. To locate change, the most intuitive way is to subtract two images \cite{park2019robust,hosseinzadeh2021image}, but this risks computing difference features with noise if two images are unaligned \cite{tu-etal-2021-r}. Recently, researchers \cite{shi2020finding} find that same objects from different viewpoints would have similar features, so they match object features between two images to predict difference features. This paradigm has been followed by some of the recent works \cite{kim2021agnostic,Qiu_2021_ICCV,yao2022image,tu-etal-2021-r,tu2023neighoborhood}. 

Despite the progress,  current match-based methods suffer from learning stable difference features under pseudo changes. In detail, the matching is directly modeled between two image features, usually by cross-attention.  However, the features of corresponding objects might shift under pseudo change. This case is more severe under drastic viewpoint changes (Fig. \ref{fig1} (d)). Such feature shift appearing in most objects would overwhelm the local feature change, thus making it less effective to directly match two images. 

For this challenge, we have two new observations.   (1)  While the feature difference might be ignored between a pair of similar images, it is hard to be overwhelmed between two images from different pairs.  As such, contrastive difference learning between similar/dissimilar images can help the model focus more on the change of feature and resist feature shift.  (2) Pseudo changes are essentially different distortions of objects, so they just construct cross-view comparison between two similar images, rather than affecting their similarity.  Motivated by these, we study cross-view feature matching between similar/dissimilar images, and maximize the alignment of similar ones, so as to learn two view-invariant image representations. Based on these, we can reconstruct the representations of unchanged objects and learn a stable difference representation.


In this paper, we tackle the above challenge with a novel \textbf{S}elf-supervised \textbf{C}r\textbf{O}ss-view \textbf{RE}presentation \textbf{R}econstruction (SCORER) network, which learns a stable difference representation while resisting pseudo changes for caption generation. Concretely, given two similar images,  we first devise a multi-head token-wise matching (MTM) to model relationships between cross-view features from similar/dissimilar images, via fully interacting different feature subspaces. Then, by maximizing cross-view contrastive alignment of the given image pair, SCORER learns their representations that are invariant to pseudo changes in a self-supervised way. Based on these, SCORER mines their reliable common features by cross-attention, so as to reconstruct the representations of unchanged objects.  Next, we fuse the representations into two images to highlight the unchanged objects and implicitly infer the difference. Through this manner, we can obtain the difference representation that not only captures the change, but also conserves referent information, thus generating a high-level linguistic sentence with a transformer decoder.

To improve the quality of sentence,  we further design a cross-modal backward reasoning (CBR) module. CBR first reversely produces a ``hallucination'' representation with the full representations of sentence and ``before'' image, where the ``hallucination'' is modeled based on the viewpoint of ``before''. Then, we push it closer to the ``after'' representation by maximizing their cross-view contrastive alignment. Through this self-supervised manner, we ensure that the generated sentence is informative about the difference.



\textbf{Our key contributions are}: \textbf{(1)} We propose SCORER to learn two view-invariant image representations for reconstructing the representations of unchanged objects, so as to model a stable difference representation under pseudo changes. 
 \textbf{(2)} We devise MTM to model relationships between cross-view images by fully interacting their different feature subspaces,  which plays a critical role in view-invariant representation learning. \textbf{(3)}  We design CBR to improve captioning quality by enforcing the generated caption is informative about the difference. \textbf{(4)} Our method performs favorably against  the state-of-the-art methods on four public datasets with different change scenarios.

\begin{figure*}[t]
\centering
\includegraphics[width=1\textwidth]{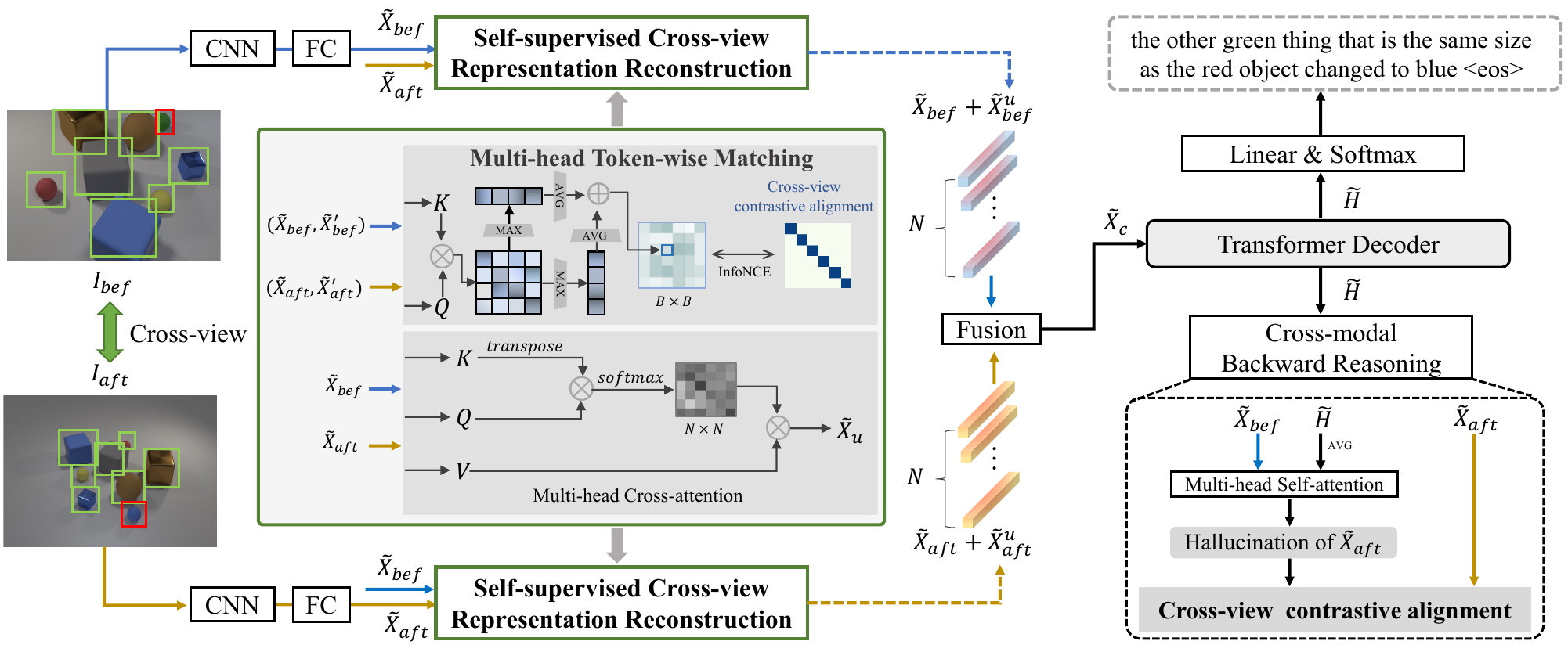} 
  \caption{The architecture of the proposed method, including a pre-trained CNN model,  the \textbf{self-supervised cross-view representation reconstruction} network, a transformer decoder, and  the \textbf{cross-modal backward reasoning} module. $\tilde {X}'_{bef}$ and $\tilde {X}'_{aft}$ denote the ``before'' and ``after'' image features from different pairs in the training batch. $B$ is the batch size; $N$ indicates the feature number in each image.   }
\label{fig2}
\end{figure*}

\section{Related Work}
\label{related work}
\textbf{Change Captioning }  is a new task in vision-language understanding and generation \cite{li2022long,liu2022entity,lin2022swinbert,tu2022i2transformer,cong2022ls,wang2023semantic}. The pioneer works \cite{jhamtani2018learning,tan2019expressing} describe the difference between two aligned images (Fig. \ref{fig1} (a) (b)). Since there usually exist viewpoint changes in a dynamic environment, recent works \cite{park2019robust,kim2021agnostic} collect two datasets to simulate moderate (Fig. \ref{fig1}  (c)) and extreme viewpoint changes (Fig. \ref{fig1}  (d)). To describe the difference under viewpoint changes, previous works \cite{park2019robust,liao2021scene} compute the difference by direct subtraction, which could compute difference with noise \cite{shi2020finding}. Recent methods \cite{kim2021agnostic,Qiu_2021_ICCV,tu-etal-2021-r,tu2023neighoborhood,tu2023viewpoint,yue2023i3n} directly match two images to predict difference features. However, due to the influence of pseudo changes, these methods are hard to learn stable difference features. In contrast, our SCORER first learns two view-invariant image representations by maximizing their cross-view contrastive alignment. Then, it mines their common features to reconstruct the representations of unchanged objects, thus learning a stable difference representation for caption generation.
We note that the latest work \cite{yao2022image} pre-trains the model with three self-supervised tasks, in order to improve cross-modal alignment. Different from it, we enforce the cross-modal alignment by implementing cross-modal backward reasoning in a self-supervised way. Meanwhile, our overall architecture is trained in an end-to-end manner, which improves the training efficiency.

\textbf{Token-wise Matching} has been used in latest image/video retrieval works \cite{yao2021filip,wang2022disentangled} to compute cross-modal interaction between image/video and text features. However, since pseudo changes would induce feature shift between object pairs, it is insufficient to only match cross-view features at token level. Hence, we further design a multi-head token-wise matching for finer-level interaction between different feature subspaces of cross-view images. This is key to learn the view-invariant representations. 
 
\textbf{Cross-modal Consistency Constraint } is to verify the quality of caption by using it and ``before'' image to rebuild ``after'' image. This idea has been tried by the latest works \cite{hosseinzadeh2021image,kim2021agnostic}. However, both works only enforce the consistency among the caption, the changed object in ``before'' and ``after'' images, while ignoring the constraint for referents. If the changed object is similar to other objects (Fig. \ref{fig1} (c) (d)), describing both the change and its referent is helpful to convey accurate change information.  Considering this, we perform backward reasoning with the full representations of ``before'' and ``after'' images, which helps generate a high-level sentence about the change and its referent.

\section{Methodology}
As shown in Fig. \ref{fig2}, our method consists of four parts: (1) A pre-trained CNN encodes a pair of cross-view images into two representations. (2) The proposed SCORER learns two view-invariant representations to reconstruct the representations of unchanged objects and model the difference representation. (3) A transformer decoder translates the difference representation into a high-level linguistic sentence. (4) The proposed CBR improves the quality of sentence via enforcing  it to be informative about the difference.

\subsection{Cross-view Image Pair Encoding}
Formally, given a pair of images ``before'' $I_{bef}$ and ``after'' $I_{aft}$, we utilize a pre-trained CNN model to extract their grid features, denoted as $X_{bef}$ and $X_{aft}$, where $X \in \mathbb{R}^{C \times H \times W}$. $C$, $H$, $W$ indicate the number of channels, height, and width. We first project both representations into a low-dimensional embedding space of $\mathbb{R}^{D}$: 
\begin{equation}
\tilde {X}_{o}=\text{conv}_2 ({X}_{o})  + pos({X}_{o}),\\
\end{equation}
where $o \in (bef,aft)$. $\text{conv}_2$ denotes a 2D-convolutional layer; $pos$ is a learnable position embedding layer.

\subsection{Self-supervised Cross-view Representation Reconstruction}
The core module of SCORER is the multi-head token wise matching (MTM). MTM aims to model relationships between cross-view images by performing fine-grained interaction between different feature subspaces, which plays a key role in view-invariant representation learning. In the following, we first elaborate MTM and then describe how to use it for view-invariant representation learning. Finally, we introduce how to reconstruct the representations of unchanged objects  for difference representation learning.

\subsubsection{Multi-head Token-wise Matching.}  We first introduce the single-head token-wise matching (TM) and then extend it into the multi-head version. Formally, given a query $Q \in \mathbb{R}^{N \times D}$ and a key $K \in \mathbb{R}^{N \times  D}$,  we first compute the similarity of $i$-th query token with all key tokens and select the maximum one as its token-wise maximum similarity with $K$. Then, we perform average pooling over the token-wise maximum similarity of all query tokens to obtain the similarity of $Q$ to $K$. By analogy, we compute the average token-wise maximum similarity of $K$ to $Q$, which ensures capturing correct relationships between them.  The above computation is formulated as follows:

\begin{equation}
\begin{gathered}
\text{TM}(Q, K)=\left[\frac{1}{N} \sum_{i=1}^N \max _{j=1}^N\left(e_{i, j}\right)+\frac{1}{N} \sum_{j=1}^N \max _{i=1}^N\left(e_{i, j}\right)\right] / 2, \\
e_{i, j}=\left(q_i\right)^\top k_j.
\end{gathered}
\end{equation}
Further, we extend $\text{TM}$ into a multi-head version to jointly match different feature subspaces of $Q$ and $K$, so as to perform fine-grained interaction between them:
\begin{equation}
\begin{gathered}
\text{MTM}(Q, K)=\text { Concat }_{i'=1 \ldots h}\left(\text { head }_{i'}\right), \\
\text { head }_{i'}=\text{TM}\left(Q W_{i'}^Q, K W_{i'}^K\right).
\end{gathered}
\end{equation}

\subsubsection{View-invariant Representation Learning}
 In a training batch, we sample $B$ image pairs of ``before'' and ``after''. For $k$-th ``before'' image  $\tilde{X}_k^{b}$,  $k$-th ``after'' image $\tilde{X}_k^{a}$ is its positive, while other ``after'' images  will be the negatives in this batch.  First, we reshape $\tilde X \in \mathbb{R}^{D \times H \times W}$ to $\tilde X\in\mathbb{R}^{N \times D}$, where $N=HW$ denotes the number of features. Then, we use MTM to compute similarity ($B \times B$  matrix) of ``before'' to ``after'' and ``after'' to ``before'', respectively. Next, we maximize cross-view contrastive alignment between $\tilde{X}_k^{b}$ and $\tilde{X}_k^{a}$ while minimizing the alignment of non-similar images, by the InfoNCE loss \cite{oord2018representation}:

\begin{equation}
\label{infonce}
\begin{gathered}
\mathcal{L}_{b 2 a}=-\frac{1}{B} \sum_k^B \log \frac{\exp \left(\text{MTM}\left(\tilde{X}_k^{b}, \tilde{X}_k^{a}\right) / \tau\right)}{\sum_r^B \exp \left(\text{MTM}\left(\tilde{X}_k^{b}, \tilde{X}_r^{a}\right) / \tau\right)}, \\
\mathcal{L}_{a 2 b}=-\frac{1}{B} \sum_k^B \log \frac{\exp \left(\text{MTM}\left(\tilde{X}_k^{a}, \tilde{X}_k^{b}\right) / \tau\right)}{\sum_r^B \exp \left(\text{MTM}\left(\tilde{X}_k^{a},
\tilde{X}_r^{b}\right) / \tau\right)}, \\
\mathcal{L}_{\text {cv }}=\frac{1}{2}(\mathcal{L}_{b 2 a}+\mathcal{L}_{a 2 b}),
\end{gathered}
\end{equation}
where $\tau$  is the temperature hyper-parameter.  In this self-supervised way, we can make the representations of $\tilde X_{bef}$ and $\tilde X_{aft}$ invariant to pseudo changes, so as to facilitate the following cross-view representation reconstruction.

\subsubsection{Cross-view Representation Reconstruction}
Based on the two view-invariant representations $\tilde X_{bef}$ and $\tilde X_{aft}$, we use a multi-head cross-attention (MHCA) \cite{vaswani2017attention} to mine their common features for reconstructing the representations of unchanged objects in each image. Here, representation reconstruction indicates that the unchanged representations of each image are distilled from the other one,  \textit{e.g.}, the unchanged representations of $\tilde X_{bef}$ are computed by transferring similar features on $\tilde X_{aft}$ back to the corresponding positions on $\tilde X_{bef}$. In this way, we reconstruct the unchanged representations for each image, respectively: 
\begin{equation}
\label{mhca}
\begin{aligned}
\tilde X_{bef}^u=\text {MHCA }(\tilde X_{bef}, \tilde X_{aft}, \tilde X_{aft}),\\
\tilde X_{aft}^u=\text { MHCA}(\tilde X_{aft}, \tilde X_{bef}, \tilde X_{bef}).\\
\end{aligned}
\end{equation}
Then, instead of subtracting them from image representations \cite{shi2020finding,tu-etal-2021-r,tu2023neighoborhood}, which leads to information (\emph{e.g.,} referents) loss, we integrate them into image representations to highlight the unchanged objects and deduce the difference information, so as to learn the stable difference representation in each image:
\begin{equation}
\tilde X_{o}^c= \text{LN}(\tilde X_{o}+\tilde X_{o}^u). \\
\end{equation}
Herein, $o \in (bef, aft)$ and LN is short for LayerNorm \cite{ba2016layer}. Finally, we obtain the difference representation between two images by fusing $\tilde X_{bef}^c$ and $\tilde X_{aft}^c$, which is implemented by a fully-connected layer with the ReLU function:
\begin{equation}
\tilde{X}_{c}=\operatorname{ReLU}\left(\left[\tilde{X}_{bef}^{c} ; \tilde{X}_{aft}^{c}\right] W_{h} + b_h \right ),
\end{equation}
where [;] is a concatenation operation.

\subsection{Caption Generation}
After leaning $ \tilde{X}_{c} \in \mathbb{R}^{N \times D}$, we use a transformer decoder \cite{vaswani2017attention} to translate it into a sentence.  First, the multi-head self-attention takes the word  features $ E[W]=\{E[w_1],... ,E[w_{m}]\}$ (ground-truth words during training, predicted words during inference) as inputs and computes a set of intra-relation embedded word features, denoted as $\hat E[W]$.
Then, the decoder utilizes $ \hat E[W]$ to query the most related features $\hat H$ from $ \tilde{X}_{c} $ via the multi-head cross-attention.
Afterward, the $\hat H$ is passed to a feed-forward network to obtain an enhanced representation $\tilde H$.
Finally, the probability distributions of target words are calculated by:
\begin{equation}
\label{word}
\tilde W=\operatorname{Softmax}\left(\tilde H W_{c}+{b}_{c}\right),
\end{equation}
where $W_{c}\in \mathbb{R}^{D \times U}$ and $b_{c} \in \mathbb{R}^{U}$ are the parameters to be learned; $U$ is the dimension of vocabulary size.

\subsection{Cross-modal Backward Reasoning}
To improve the quality of generated sentence, we devise the CBR to first reversely model a ``hallucination'' representation with the sentence and ``before'' image. Then, we push it closer to the  ``after'' representation to enforce the sentence to be informative about the difference. Concretely, we first fuse $\tilde H \in \mathbb{R}^{m \times D}$ by the mean-pooling operation to obtain a sentence feature $\tilde T$. Then, we broadcast $\tilde T \in \mathbb{R}^{D}$ as $\tilde T \in \mathbb{R}^{D \times H \times W}$ and  concatenate it with $\tilde X_{bef}$, so as to obtain the ``hallucination'' $\hat{X}_{hal}$:
\begin{equation}
\hat{X}_{hal}=\text{conv}_2([\tilde X_{bef}; \tilde T ]), \hat{X}_{hal} \in \mathbb{R}^{D \times H \times W}.
\end{equation}
 $\hat{X}_{hal}$ and $\tilde X_{bef}$ are kept as the same shape to ensure that spatial information is not collapsed. Next, we capture the relationships between different locations in $\hat{X}_{hal}$ based on the multi-head self-attention (MHSA), which is essential for backward reasoning and computed by:
\begin{equation}
\tilde{X}_{hal}=\text{conv}_2[\text {MHSA }(\hat{X}_{hal}, \hat{X}_{hal}, \hat{X}_{hal})],
\end{equation}
Since the ``hallucination'' representation is produced based on the viewpoint of ``before'' representation, it is less effective to directly match it with the ``after'' representation. 

To this end, we sample unrelated representations of ``hallucination'' and ``after'' from different pairs, which are as erroneous candidates for CBR. Similarly, in each batch, for $k$-th ``hallucination'' $\tilde {X}_k^{h}$,  $k$-th ``after'' $\tilde{X}_k^{a}$ is its positive, while the other ``after'' images  will be the negatives. Also, we use MTM to capture relationships  between positive/negative pairs. Subsequently, we maximize cross-view contrastive alignment of positive pairs by the InfoNCE loss \cite{oord2018representation}, which is similar to Eq. (\ref{infonce}):  
\begin{equation}
\mathcal{L}_{\text {cm}}=\frac{1}{2}(\mathcal{L}_{h 2 a}+\mathcal{L}_{a 2 h}).
\end{equation}
Through this self-supervised manner, we make the sentence sufficiently describe the difference information.

\subsection{Joint Training}
The proposed overall network is trained in an end-to-end manner by maximizing the likelihood of the observed  word sequence. Given the ground-truth words $\left(w_{1}^{*}, \ldots, w_{m}^{*}\right)$, we minimize the negative log-likelihood loss:
\begin{equation}
\mathcal L_{cap}(\theta)=-\sum_{t=1}^{m} \log p_\theta \left(w_{t}^{*} \mid w_{<t}^{*}\right),
\end{equation}
where $p_\theta\left(w_t^{*} \mid w_{<t}^{*}\right)$ is computed by Eq.~(\ref{word}), and $\theta$ are the parameters of the network. Besides,  the network is self-supervised by the losses of two contrastive alignments. Hence, the final loss function is optimized as follows:
\begin{equation}
\label{cross-entropy}
\mathcal L =\mathcal L_{cap} + \lambda_v \mathcal{L}_{\text {cv}} + \lambda_m \mathcal{L}_{\text {cm}},
\end{equation}
where $\lambda_{v}$ and $\lambda_{m}$ are the trade-off parameters, which are discussed in the supplementary material. 

\section{Experiments}
\label{experiment}
\subsection{Datasets}
\textbf{CLEVR-Change} is a large-scale dataset \cite{park2019robust} with moderate viewpoint change. It has 79,606 image pairs, including five change types, \emph{i.e.}, ``Color'', ``Texture'', ``Add'', ``Drop'',  and ``Move''.We use the official split with 67,660 for training, 3,976 for validation and 7,970 for testing.

\textbf{CLEVR-DC} is a large-scale dataset \cite{kim2021agnostic} with extreme viewpoint shift. It includes 48,000 pairs with same change types as CLEVR-Change. We use the official split with  85\% for training, 5\% for validation, and 10\% for testing.

\textbf{Image Editing Request} dataset \cite{tan2019expressing} includes 3,939 aligned image pairs with 5,695 editing  instructions.  We use the official split with 3,061 image pairs for training, 383 for validation, and 495 for testing.

\textbf{Spot-the-Diff} dataset \cite{jhamtani2018learning} includes 13,192 aligned image pairs from surveillance cameras. Following SOTA methods, we mainly evaluate our model in a single change setting.  Based on the official split, the dataset is split into training, validation, and testing with a ratio of 8:1:1.

\subsection{Evaluation Metrics}
Following the current state-of-the-art methods, five metrics are used to evaluate the generated sentences, \emph{i.e.}, BLEU-4 (B) \cite{papineni2002bleu}, METEOR (M) \cite{banerjee2005meteor}, ROUGE-L (R) \cite{lin2004rouge}, CIDEr (C) \cite{vedantam2015cider}, and SPICE (S) \cite{anderson2016spice}. The results are computed based on the Microsoft COCO evaluation server \cite{chen2015microsoft}.

\subsection{Implementation Details}
For a  fair comparison, we follow the SOTA methods to use a pre-trained ResNet-101  \cite{he2016deep} to extract grid features of an image pair, with the dimension of 1024 $\times$ 14 $\times$ 14. We first project these  features into a lower dimension of 512. The hidden size in the overall model and word embedding size are set to 512 and 300.  The proper head and layer numbers of SCORER are discussed below. The head and layer numbers in the decoder are set to 8 and 2 on the four  datasets. 
During training, We use Adam optimizer \cite{kingma2014adam} to minimize the negative log-likelihood loss of Eq. (\ref{cross-entropy}). During inference, the greedy decoding strategy is used to generate captions. Both training and inference are implemented with PyTorch \cite{paszke2019pytorch} on an RTX 3090 GPU. More implementation details are described in the supplementary material.

\subsection{Performance Comparison}
\subsubsection{Results on the CLEVR-Change Dataset.}
We compare with the state-of-the-art methods in: 1) total performance under both semantic and pseudo changes; 2) semantic change; 3) different change types. The comparison methods are categorized into 1) end-to-end training: DUDA \cite{park2019robust}, DUDA+ \cite{hosseinzadeh2021image}, R$^{3}$Net+SSP \cite{tu-etal-2021-r}, VACC \cite{kim2021agnostic}, SRDRL+AVS \cite{tu2021semantic}, SGCC \cite{liao2021scene},  MCCFormers-D \cite{Qiu_2021_ICCV}, IFDC \cite{huang2022image}, BDLSCR \cite{sun2022bidirectional}, NCT \cite{tu2023neighoborhood}, and VARD-Trans \cite{tu2023viewpoint}; 2) reinforcement learning: M-VAM+RAF \cite{shi2020finding}; 3) pre-training:  PCL w/ pre-training \cite{yao2022image}. 

\begin{table*}[h]
  \centering
 
   \begin{center}
    \begin{tabular}{c|ccccc|ccccc}
    \toprule
          & \multicolumn{5}{c|}{Total}            & \multicolumn{5}{c}{Semantic Change} \\
    Method & B     & M     & R     & C     & S     & B     & M     & R     & C     & S \\
    \midrule
    PCL w/ Pre-training (AAAI 2022) \cite{yao2022image} & \cellcolor[rgb]{ .878,  .953,  .984}51.2 & \cellcolor[rgb]{ .878,  .953,  .984}36.2 & \cellcolor[rgb]{ .867,  .922,  .969}71.7 & \cellcolor[rgb]{ .608,  .761,  .902}\textbf{128.9} & -     & -     & -     & -     & -     & - \\
    \midrule
    M-VAM+RAF (ECCV 2020) \cite{shi2020finding} & \cellcolor[rgb]{ .878,  .953,  .984}51.3 & \cellcolor[rgb]{ .867,  .922,  .969}37.8 & \cellcolor[rgb]{ .867,  .922,  .969}70.4 & \multicolumn{1}{c}{\cellcolor[rgb]{ .878,  .953,  .984}115.8} & \cellcolor[rgb]{ .867,  .922,  .969}30.7 & -     & -     & -     & -     & - \\
    \midrule
    DUDA (ICCV 2019) \cite{park2019robust} & \cellcolor[rgb]{ .878,  .953,  .984}47.3 & \cellcolor[rgb]{ .878,  .953,  .984}33.9 & -     & \cellcolor[rgb]{ .878,  .953,  .984}112.3 & \cellcolor[rgb]{ .878,  .953,  .984}24.5 & 42.9  & 29.7  & -     & 94.6  & 19.9 \\
    DUDA+ (CVPR 2021) \cite{hosseinzadeh2021image} & \cellcolor[rgb]{ .878,  .953,  .984}51.2 & \cellcolor[rgb]{ .867,  .922,  .969}37.7 & \cellcolor[rgb]{ .867,  .922,  .969}70.5 & \cellcolor[rgb]{ .878,  .953,  .984}115.4 & \cellcolor[rgb]{ .867,  .922,  .969}31.1 & \cellcolor[rgb]{ .878,  .953,  .984}49.9 & \cellcolor[rgb]{ .878,  .953,  .984}34.3 & \cellcolor[rgb]{ .878,  .953,  .984}65.4 & \cellcolor[rgb]{ .878,  .953,  .984}101.3 & \cellcolor[rgb]{ .878,  .953,  .984}27.9 \\
    R$^{3}$Net+SSP (EMNLP 2021) \cite{tu-etal-2021-r} & \cellcolor[rgb]{ .867,  .922,  .969}54.7 & \cellcolor[rgb]{ .867,  .922,  .969}39.8 & \cellcolor[rgb]{ .867,  .922,  .969}73.1 & \cellcolor[rgb]{ .867,  .922,  .969}123.0 & \cellcolor[rgb]{ .867,  .922,  .969}32.6 & \cellcolor[rgb]{ .867,  .922,  .969}52.7 & \cellcolor[rgb]{ .867,  .922,  .969}36.2 & \cellcolor[rgb]{ .867,  .922,  .969}69.8 & \cellcolor[rgb]{ .867,  .922,  .969}116.6 & \cellcolor[rgb]{ .867,  .922,  .969}30.3 \\
    VACC (ICCV 2021) \cite{kim2021agnostic} & \cellcolor[rgb]{ .878,  .953,  .984}52.4 & \cellcolor[rgb]{ .867,  .922,  .969}37.5 & -     & \cellcolor[rgb]{ .878,  .953,  .984}114.2 & \cellcolor[rgb]{ .867,  .922,  .969}31.0 & -     & -     & -     & -     & - \\
     SGCC (ACM MM 2021) \cite{liao2021scene} & \cellcolor[rgb]{ .878,  .953,  .984}51.1 & \cellcolor[rgb]{ .741,  .843,  .933}40.6 & \cellcolor[rgb]{ .867,  .922,  .969} 73.9     & \cellcolor[rgb]{ .867,  .922,  .969}121.8 & \cellcolor[rgb]{ .867,  .922,  .969}32.2 & -     & -     & -     & -     & - \\
    SRDRL+AVS (ACL 2021) \cite{tu2021semantic} & \cellcolor[rgb]{ .867,  .922,  .969}54.9 & \cellcolor[rgb]{ .741,  .843,  .933}40.2 & \cellcolor[rgb]{ .867,  .922,  .969}73.3 & \cellcolor[rgb]{ .867,  .922,  .969}122.2 & \cellcolor[rgb]{ .741,  .843,  .933}32.9 & \cellcolor[rgb]{ .867,  .922,  .969}52.7 & \cellcolor[rgb]{ .867,  .922,  .969}36.4 & \cellcolor[rgb]{ .867,  .922,  .969}69.7 & \cellcolor[rgb]{ .867,  .922,  .969}114.2 & \cellcolor[rgb]{ .867,  .922,  .969}30.8 \\
    MCCFormers-D (ICCV 2021) \cite{Qiu_2021_ICCV} & \cellcolor[rgb]{ .878,  .953,  .984}52.4 & \cellcolor[rgb]{ .867,  .922,  .969}38.3 & -     & \cellcolor[rgb]{ .867,  .922,  .969}121.6 & \cellcolor[rgb]{ .878,  .953,  .984}26.8 & -     & -     & -     & -     & - \\
    IFDC (TMM 2022) \cite{huang2022image} & \cellcolor[rgb]{ .878,  .953,  .984}49.2 & \cellcolor[rgb]{ .878,  .953,  .984}32.5 & \cellcolor[rgb]{ .878,  .953,  .984}69.1 & \cellcolor[rgb]{ .878,  .953,  .984}118.7 & -     & \cellcolor[rgb]{ .878,  .953,  .984}47.2 & 29.3  & \cellcolor[rgb]{ .878,  .953,  .984}63.7 & \cellcolor[rgb]{ .878,  .953,  .984}105.4 & - \\
    
    NCT (TMM 2023) \cite{tu2023neighoborhood} & \cellcolor[rgb]{ .741,  .843,  .933}55.1 & \cellcolor[rgb]{ .741,  .843,  .933}40.2 & \cellcolor[rgb]{ .741,  .843,  .933}73.8 & \cellcolor[rgb]{ .867,  .922,  .969}124.1 & \cellcolor[rgb]{ .741,  .843,  .933}32.9 & \cellcolor[rgb]{ .867,  .922,  .969}53.1 & \cellcolor[rgb]{ .867,  .922,  .969}36.5 & \cellcolor[rgb]{ .867,  .922,  .969}70.7 & \cellcolor[rgb]{ .867,  .922,  .969}118.4 & \cellcolor[rgb]{ .741,  .843,  .933}30.9 \\
    VARD-Trans (TIP 2023) \cite{tu2023viewpoint} & \cellcolor[rgb]{ .741,  .843,  .933}55.4 & \cellcolor[rgb]{ .741,  .843,  .933}40.1 & \cellcolor[rgb]{ .741,  .843,  .933}73.8 & \cellcolor[rgb]{ .741,  .843,  .933}126.4 & \cellcolor[rgb]{ .741,  .843,  .933}32.6  & -     & -     & -     & -     & - \\
    \textbf{SCORER (Ours)} & \cellcolor[rgb]{ .741,  .843,  .933}55.8 & \cellcolor[rgb]{ .741,  .843,  .933}40.8 & \cellcolor[rgb]{ .741,  .843,  .933}74.0 & \cellcolor[rgb]{ .741,  .843,  .933}126.0 & \cellcolor[rgb]{ .741,  .843,  .933}33.0 & \cellcolor[rgb]{ .741,  .843,  .933}54.1 & \cellcolor[rgb]{ .741,  .843,  .933}37.4 & \cellcolor[rgb]{ .741,  .843,  .933}71.5 & \cellcolor[rgb]{ .741,  .843,  .933}122.0 & \cellcolor[rgb]{ .741,  .843,  .933}31.2 \\
    \textbf{SCORER+CBR (Ours)} & \cellcolor[rgb]{ .608,  .761,  .902}\textbf{56.3} & \cellcolor[rgb]{ .608,  .761,  .902}\textbf{41.2} & \cellcolor[rgb]{ .608,  .761,  .902}\textbf{74.5} & \cellcolor[rgb]{ .608,  .761,  .902}\textbf{126.8} & \cellcolor[rgb]{ .608,  .761,  .902}\textbf{33.3} & \cellcolor[rgb]{ .608,  .761,  .902}\textbf{54.4} & \cellcolor[rgb]{ .608,  .761,  .902}\textbf{37.6} & \cellcolor[rgb]{ .608,  .761,  .902}\textbf{71.7} & \cellcolor[rgb]{ .608,  .761,  .902}\textbf{122.4} & \cellcolor[rgb]{ .608,  .761,  .902}\textbf{31.6} \\
    \bottomrule
    \end{tabular}%
  \end{center}
  
   \caption{Comparison with the state-of-the-art methods on CLEVR-Change under the settings of total performance and  semantic change.}
   \label{total_com}%
\end{table*}%

\begin{table}[t]
   \small
    \begin{center}
    
   \begin{tabular}{c|ccccc}
    \toprule
          & \multicolumn{5}{c}{CIDEr} \\
    Method & CL     & T     & A     & D     & MV \\
    \midrule
    PCL w/ PT & \cellcolor[rgb]{ .886,  .937,  .855}131.2 & \cellcolor[rgb]{ .886,  .937,  .855}101.1 & \cellcolor[rgb]{ .663,  .816,  .557}\textbf{133.3} & \cellcolor[rgb]{ .953,  .976,  .929}116.5 & \cellcolor[rgb]{ .886,  .937,  .855}81.7 \\
    \midrule
    M-VAM+RAF  & \cellcolor[rgb]{ .953,  .976,  .929}122.1 & \cellcolor[rgb]{ .886,  .937,  .855}98.7 & \cellcolor[rgb]{ .886,  .937,  .855}126.3 & \cellcolor[rgb]{ .953,  .976,  .929}115.8 & \cellcolor[rgb]{ .886,  .937,  .855}82.0 \\
    DUDA  & \cellcolor[rgb]{ .953,  .976,  .929}120.4 & \cellcolor[rgb]{ .953,  .976,  .929}86.7 & \cellcolor[rgb]{ .953,  .976,  .929}108.2 & \cellcolor[rgb]{ .953,  .976,  .929}103.4 & 56.4 \\
    DUDA+ & \cellcolor[rgb]{ .953,  .976,  .929}120.8 & \cellcolor[rgb]{ .953,  .976,  .929}89.9 & \cellcolor[rgb]{ .886,  .937,  .855}119.8 & \cellcolor[rgb]{ .953,  .976,  .929}123.4 & \cellcolor[rgb]{ .953,  .976,  .929}62.1 \\
    R$^{3}$Net+SSP  & \cellcolor[rgb]{ .886,  .937,  .855}139.2 & \cellcolor[rgb]{ .886,  .937,  .855}123.5 & \cellcolor[rgb]{ .886,  .937,  .855}122.7 & \cellcolor[rgb]{ .953,  .976,  .929}121.9 & \cellcolor[rgb]{ .886,  .937,  .855}88.1 \\
    SRDRL+AVS  & \cellcolor[rgb]{ .886,  .937,  .855}136.1 & \cellcolor[rgb]{ .886,  .937,  .855}122.7 & \cellcolor[rgb]{ .886,  .937,  .855}121.0 & \cellcolor[rgb]{ .886,  .937,  .855}126.0 & \cellcolor[rgb]{ .886,  .937,  .855}78.9 \\
    BDLSCR  & \cellcolor[rgb]{ .886,  .937,  .855}136.1 & \cellcolor[rgb]{ .886,  .937,  .855}122.7 & \cellcolor[rgb]{ .886,  .937,  .855}121.0 & \cellcolor[rgb]{ .886,  .937,  .855}126.0 & \cellcolor[rgb]{ .886,  .937,  .855}78.9 \\
    IFDC  & \cellcolor[rgb]{ .886,  .937,  .855}133.2 & \cellcolor[rgb]{ .886,  .937,  .855}99.1 & \cellcolor[rgb]{ .886,  .937,  .855}128.2 & \cellcolor[rgb]{ .953,  .976,  .929}118.5 & \cellcolor[rgb]{ .886,  .937,  .855}82.1 \\
    NCT   & \cellcolor[rgb]{ .886,  .937,  .855}140.2 & \cellcolor[rgb]{ .886,  .937,  .855}128.8  & \cellcolor[rgb]{ .886,  .937,  .855}128.4 & \cellcolor[rgb]{ .886,  .937,  .855}129.0 & \cellcolor[rgb]{ .886,  .937,  .855}86.0 \\
    \textbf{SCORER} & \cellcolor[rgb]{ .776,  .878,  .706}143.2 & \cellcolor[rgb]{ .776,  .878,  .706}135.2 & \cellcolor[rgb]{ .776,  .878,  .706}129.4 & \cellcolor[rgb]{ .776,  .878,  .706}132.6 & \cellcolor[rgb]{ .776,  .878,  .706}91.6 \\
   \textbf{ SCORER+CBR} & \cellcolor[rgb]{ .663,  .816,  .557}\textbf{146.2} & \cellcolor[rgb]{ .663,  .816,  .557}\textbf{133.7} & \cellcolor[rgb]{ .663,  .816,  .557}\textbf{131.1} & \cellcolor[rgb]{ .663,  .816,  .557}\textbf{133.9} & \cellcolor[rgb]{ .663,  .816,  .557}\textbf{92.2} \\
    \bottomrule
    \end{tabular}%
    \end{center}
     \caption{A detailed breakdown of evaluation on CLEVR-Change with different change types: ``(CL) Color'', ``(T) Textur'', ``(A) Add'', ``(D) Drop'', and ``(MV) Move''. PT is short for pre-training. } 
  \label{type_com}%
\end{table}%

In Table \ref{total_com}, our  method  achieves the best results on all metrics against the end-to-end training methods.  Besides, our method performs much better than these two methods augmented by pre-training and reinforcement learning.
We note that SCORER outperforms MCCFormers-D by a large margin.  MCCFormers-D is a classic match-based method that directly correlates two image representations to learn a difference representation, which is then fed into a transformer decoder for caption generation. Different from it, our SCORER first learns two view-invariant image representations by maximizing their cross-view contrastive alignment. Then, SCORER reconstructs the representations of unchanged objects, so as  to learn a stable difference representation under pseudo changes for caption generation.  

In Table \ref{type_com}, under the detailed change types, our method surpasses the current methods by a large margin in almost every category.  Under the most difficult type ``Move'', our SCORER+CBR achieves the relative improvement of 4.7\% against R$^{3}$Net+SSP.  This  validates the necessary of view-invariant representation learning. Moreover, under different settings, CBR helps yield an extra performance boost, which shows it does improve captioning quality.


\subsubsection{Results on the CLEVR-DC Dataset}
On CLEVR-DC with extreme viewpoint changes, we compare SCORER/SCORER+CBR with several state-of-the-art methods: DUDA/DUDA+CC \cite{park2019robust}, M-VAM/M-VAM+CC \cite{shi2020finding}, VA/VACC \cite{kim2021agnostic}, MCCFormers-D \cite{Qiu_2021_ICCV}, NCT \cite{tu2023neighoborhood}, and VARD-Trans \cite{tu2023viewpoint}. For fair-comparison, we compare them based on the usage of cross-modal consistency constraint. We implement MCCFormers-D based on the released code on CLEVR-DC and Image Editing Request datasets. 

The results are shown in Table \ref{dc_com}. Our SCORER achieves the best results on most metrics. This benefits from learning two view-invariant representations to reconstruct representations of unchanged objects, thus learning a stable difference representation under extreme viewpoint changes. When we implement CBR, the performance of SCORER+CBR is further boosted, especially achieving 16.7\% improvement against VACC on CIDEr. This shows that our CBR  can calibrate the model to generate a linguistic sentence describing the change and its referent.

\begin{table}[t]
  \centering
 \begin{center}
   
    \begin{tabular}{c|cccc}
    \toprule
    Method & B     & M     & C     & S \\
    \midrule
    DUDA \cite{park2019robust} & \cellcolor[rgb]{ .855,  .91,  .973}40.3 & \cellcolor[rgb]{ .855,  .91,  .973}27.1 & 56.7  & \cellcolor[rgb]{ .706,  .776,  .906}16.1 \\
    M-VAM \cite{shi2020finding} & \cellcolor[rgb]{ .855,  .91,  .973}40.9 & \cellcolor[rgb]{ .855,  .91,  .973}27.1 & \cellcolor[rgb]{ .855,  .91,  .973}60.1 & \cellcolor[rgb]{ .706,  .776,  .906}15.8 \\
    VA \cite{kim2021agnostic}   & \cellcolor[rgb]{ .851,  .882,  .949}44.5 & \cellcolor[rgb]{ .851,  .882,  .949}29.2 & \cellcolor[rgb]{ .851,  .882,  .949}70.0 & \cellcolor[rgb]{ .557,  .663,  .859}\textbf{17.1} \\
    MCCFormers-D \cite{Qiu_2021_ICCV} & \cellcolor[rgb]{ .706,  .776,  .906}46.9 & \cellcolor[rgb]{ .706,  .776,  .906}31.7 & \cellcolor[rgb]{ .851,  .882,  .949}71.6 & \cellcolor[rgb]{ .851,  .882,  .949}14.6 \\
    NCT \cite{tu2023neighoborhood}  & \cellcolor[rgb]{ .706,  .776,  .906}47.5 & \cellcolor[rgb]{ .706,  .776,  .906}32.5 & \cellcolor[rgb]{ .706,  .776,  .906}76.9 & \cellcolor[rgb]{ .706,  .776,  .906}15.6 \\
    VARD-Trans \cite{tu2023viewpoint}  & \cellcolor[rgb]{ .706,  .776,  .906}48.3 & \cellcolor[rgb]{ .706,  .776,  .906}32.4 & \cellcolor[rgb]{ .706,  .776,  .906}77.6 & \cellcolor[rgb]{ .706,  .776,  .906}15.4\\
    \textbf{SCORER} & \cellcolor[rgb]{ .557,  .663,  .859}\textbf{49.5} & \cellcolor[rgb]{ .557,  .663,  .859}\textbf{33.4} & \cellcolor[rgb]{ .557,  .663,  .859}\textbf{82.4} & \cellcolor[rgb]{ .706,  .776,  .906}15.8 \\
    \midrule
    DUDA+CC  \cite{park2019robust} & \cellcolor[rgb]{ .851,  .882,  .949}41.7 & \cellcolor[rgb]{ .851,  .882,  .949}27.5 & \cellcolor[rgb]{ .855,  .91,  .973}62.0 & \cellcolor[rgb]{ .706,  .776,  .906}16.4 \\
    M-VAM+CC \cite{shi2020finding} & \cellcolor[rgb]{ .851,  .882,  .949}41.0 & \cellcolor[rgb]{ .851,  .882,  .949}27.2 & \cellcolor[rgb]{ .855,  .91,  .973}62.0 & \cellcolor[rgb]{ .851,  .882,  .949}15.7 \\
    VACC  \cite{kim2021agnostic} & \cellcolor[rgb]{ .706,  .776,  .906}45.0 & \cellcolor[rgb]{ .706,  .776,  .906}29.3 & \cellcolor[rgb]{ .706,  .776,  .906}71.7 & \cellcolor[rgb]{ .557,  .663,  .859}\textbf{17.6} \\
    \textbf{SCORER+CBR} & \cellcolor[rgb]{ .557,  .663,  .859}\textbf{49.4} & \cellcolor[rgb]{ .557,  .663,  .859}\textbf{33.4} & \cellcolor[rgb]{ .557,  .663,  .859}\textbf{83.7} & \cellcolor[rgb]{ .706,  .776,  .906}16.2 \\
    \bottomrule
    \end{tabular}%
    \end{center}
    \caption{Comparison with the SOTA methods on CLEVR-DC. } 
  \label{dc_com}%
\end{table}%

\subsubsection{Results on the Image Editing Reques Dataset}
To validate the generalization of our method, we conduct the experiment on a challenging dataset of Image Editing Request (IER). We  compare with the following SOTA methods: DUDA \cite{park2019robust}, Dyn rel-att \cite{tan2019expressing},  MCCFormers-D \cite{Qiu_2021_ICCV}, BDLSCR \cite{sun2022bidirectional}, NCT \cite{tu2023neighoborhood}, and VARD-Trans \cite{tu2023viewpoint}.

Table \ref{edit} shows SCORER+CBR outperforms the SOTA methods on most metrics. Especially on BLEU-4, SCORER+CBR obtains the relative improvement of 23.5\% against the latest method NCT (TMM 2023). The edited objects are usually inconspicuous. This indicates that the proposed method can fully mine the common features by maximizing cross-view contrastive alignment between two images, so as to accurately describe which part of the ``before'' image has been edited. Further, the generated sentence is refined in the process of cross-modal backward reasoning.

\begin{table}[t]
  \centering
  \begin{center}

    \begin{tabular}{c|cccc}
    \toprule
    Method & B     & M     & R     & C \\
    \midrule
    DUDA \cite{park2019robust}  & \cellcolor[rgb]{ 1,  .933,  .859}6.5 & \cellcolor[rgb]{ 1,  .933,  .859}12.4 & \cellcolor[rgb]{ 1,  .933,  .859}37.3 & 22.8 \\
    Dyn rel-att \cite{tan2019expressing} & \cellcolor[rgb]{ 1,  .933,  .859}6.7 & \cellcolor[rgb]{ 1,  .933,  .859}12.8 & \cellcolor[rgb]{ 1,  .933,  .859}37.5 & \cellcolor[rgb]{ 1,  .933,  .859}26.4 \\
    MCCFormers-D \cite{Qiu_2021_ICCV} & \cellcolor[rgb]{ .988,  .894,  .839}8.3 & \cellcolor[rgb]{ .988,  .894,  .839}14.3 & \cellcolor[rgb]{ .988,  .894,  .839}39.2 & \cellcolor[rgb]{ .988,  .894,  .839}30.2 \\
    BDLSCR \cite{sun2022bidirectional} & \cellcolor[rgb]{ 1,  .933,  .859}6.9 & \cellcolor[rgb]{ .973,  .796,  .678}14.6 & \cellcolor[rgb]{ .988,  .894,  .839}38.5 & \cellcolor[rgb]{ 1,  .933,  .859}27.7 \\
    NCT \cite{tu2023neighoborhood}  & \cellcolor[rgb]{ .988,  .894,  .839}8.1 & \cellcolor[rgb]{ .957,  .69,  .518}\textbf{15.0} & \cellcolor[rgb]{ .988,  .894,  .839}38.8 & \cellcolor[rgb]{ .973,  .796,  .678}34.2 \\
    VARD-Trans \cite{tu2023viewpoint} & \cellcolor[rgb]{ .957,  .69,  .518}\textbf{10.0} & \cellcolor[rgb]{ .973,  .796,  .678}14.8 & \cellcolor[rgb]{ .988,  .894,  .839}39.0 & \cellcolor[rgb]{ .957,  .69,  .518}\textbf{35.7} \\
    \textbf{SCORER } & \cellcolor[rgb]{ .973,  .796,  .678}9.6 & \cellcolor[rgb]{ .973,  .796,  .678}14.6 & \cellcolor[rgb]{ .973,  .796,  .678}39.5 & \cellcolor[rgb]{ .973,  .796,  .678}31.0 \\
    \textbf{SCORER+CBR  } & \cellcolor[rgb]{ .957,  .69,  .518}\textbf{10.0} & \cellcolor[rgb]{ .957,  .69,  .518}\textbf{15.0} & \cellcolor[rgb]{ .957,  .69,  .518}\textbf{39.6} & \cellcolor[rgb]{ .973,  .796,  .678}33.4 \\
    \bottomrule
    \end{tabular}%
    \end{center}
    \caption{Comparison with the SOTA methods on IER. }
  \label{edit}%
\end{table}%

\subsubsection{Results on the Spot-the-Diff Dataset}
To further validate the generalization, we conduct the experiment on Spot-the-Diff that includes aligned image pairs from the surveillance cameras. The following SOTA methods are compared: DUDA+ \cite{hosseinzadeh2021image}, M-VAM/M-VAM+RAF \cite{shi2020finding}, VACC \cite{kim2021agnostic}, SRDRL+AVS \cite{tu2021semantic},  MCCFormers-D  \cite{Qiu_2021_ICCV}, IFDC \cite{huang2022image}, BDLSCR \cite{sun2022bidirectional}, and VARD-Trans \cite{tu2023viewpoint}. 

In Table \ref{spot_com}, our method achieves  superior results on most metrics, which shows its generalization on different scenarios. Besides, our method performs lower on METEOR and SPICE when implementing CBR. Our conjecture is that image pairs on this dataset actually contain one or more changes. For fair-comparison, we conduct experiments mainly based on the single-change setup. This makes  the ``hallucination'' representation, which is reversely modeled by the ``before'' representation and single-change caption, not fully matched with the ``after'' representation. As such,  SCORER+CBR does not gain significant improvement.  

In short, compared with the state-of-the-art methods in different change scenarios, our method  achieves the impressive performance. The superiority mainly results  from that 1) SCORER learns two view-invariant image representations for reconstructing the representations of unchanged objects, so as to learn a stable difference representation for generating a linguistic sentence; 2) CBR can further improve the quality of generated sentence.

\begin{table}[t]
  \begin{center}
   \begin{tabular}{c|cccc}
    \toprule
    Method & B     & M     & C     & S \\
    \midrule
    M-VAM+RAF \cite{shi2020finding} & \cellcolor[rgb]{ 1,  .851,  .4}\textbf{11.1} & \cellcolor[rgb]{ 1,  .902,  .6}12.9 & \cellcolor[rgb]{ 1,  .851,  .4}\textbf{43.5} & \cellcolor[rgb]{ 1,  .949,  .8}17.1 \\
    \midrule
    
    M-VAM \cite{shi2020finding} & \cellcolor[rgb]{ 1,  .902,  .6}10.1 & \cellcolor[rgb]{ 1,  .902,  .6}12.4 & \cellcolor[rgb]{ 1,  .902,  .6}38.1 & 14.0 \\
    DUDA+ \cite{hosseinzadeh2021image} & \cellcolor[rgb]{ 1,  .949,  .8}8.1 & \cellcolor[rgb]{ 1,  .902,  .6}12.5 & \cellcolor[rgb]{ 1,  1,  .8}34.5 & - \\
    VACC \cite{kim2021agnostic}  & \cellcolor[rgb]{ 1,  .902,  .6}9.7 & \cellcolor[rgb]{ 1,  .902,  .6}12.6 & \cellcolor[rgb]{ 1,  .902,  .6}41.5 & - \\
    SRDRL+AVS \cite{tu2021semantic} & -     & \cellcolor[rgb]{ 1,  .902,  .6}13.0 & \cellcolor[rgb]{ 1,  1,  .8}35.3 & \cellcolor[rgb]{ 1,  .902,  .6}18.0 \\
    MCCFormers-D \cite{Qiu_2021_ICCV} & \cellcolor[rgb]{ 1,  .902,  .6}10.0 & \cellcolor[rgb]{ 1,  .902,  .6}12.4 & \cellcolor[rgb]{ 1,  .851,  .4}\textbf{43.1} & \cellcolor[rgb]{ 1,  .902,  .6}18.3 \\
    IFDC \cite{huang2022image} & \cellcolor[rgb]{ 1,  .949,  .8}8.7 & \cellcolor[rgb]{ 1,  .949,  .8}11.7 & \cellcolor[rgb]{ 1,  .949,  .8}37.0 & - \\
    BDLSCR \cite{sun2022bidirectional}  & \cellcolor[rgb]{ 1,  1,  .8}6.6 & \cellcolor[rgb]{ 1,  1,  .8}10.6 & \cellcolor[rgb]{ 1,  .902,  .6}42.2 & - \\
    VARD-Trans \cite{tu2023viewpoint} & -     & \cellcolor[rgb]{ 1,  .902,  .6}12.5 & \cellcolor[rgb]{ 1,  1,  .8}30.3 & \cellcolor[rgb]{ 1,  .949,  .8}17.3  \\
    \textbf{SCORER} & \cellcolor[rgb]{ 1,  .902,  .6}9.4 & \cellcolor[rgb]{ 1,  .851,  .4}\textbf{13.8} & \cellcolor[rgb]{ 1,  .902,  .6}38.5 & \cellcolor[rgb]{ 1,  .851,  .4}\textbf{19.3} \\
    \textbf{SCORER+CBR} & \cellcolor[rgb]{ 1,  .851,  .4}\textbf{10.2} & \cellcolor[rgb]{ 1,  .902,  .6}12.2 & \cellcolor[rgb]{ 1,  .902,  .6}38.9 & \cellcolor[rgb]{ 1,  .902,  .6}18.4 \\
    \bottomrule
    \end{tabular}%
    \end{center}
     \caption{Comparison with the SOTA methods on Spot-the-Diff.}
  \label{spot_com}%
\end{table}%

\begin{table}[t]
  \centering
  \begin{center}
   
    \begin{tabular}{c|ccccc}
    \toprule
    Ablation & B     & M     & R     & C     & S \\
    \midrule
    Subtraction & \cellcolor[rgb]{ .929,  .929,  .929}53.3 & \cellcolor[rgb]{ .929,  .929,  .929}38.8 & \cellcolor[rgb]{ .929,  .929,  .929}72.1 & \cellcolor[rgb]{ .953,  .953,  .953}119.7 & \cellcolor[rgb]{ .929,  .929,  .929}31.8 \\
    RR    & \cellcolor[rgb]{ .859,  .859,  .859}55.1 & \cellcolor[rgb]{ .859,  .859,  .859}40.5 & \cellcolor[rgb]{ .929,  .929,  .929}73.6 & \cellcolor[rgb]{ .929,  .929,  .929}123.8 & \cellcolor[rgb]{ .929,  .929,  .929}32.5 \\
    SCORER & \cellcolor[rgb]{ .859,  .859,  .859}55.8 & \cellcolor[rgb]{ .859,  .859,  .859}40.8 & \cellcolor[rgb]{ .859,  .859,  .859}74.0 & \cellcolor[rgb]{ .859,  .859,  .859}126.0 & \cellcolor[rgb]{ .859,  .859,  .859}33.0 \\
    RR+CBR & \cellcolor[rgb]{ .859,  .859,  .859}55.8 & \cellcolor[rgb]{ .859,  .859,  .859}41.0 & \cellcolor[rgb]{ .859,  .859,  .859}74.2 & \cellcolor[rgb]{ .859,  .859,  .859}125.5 & \cellcolor[rgb]{ .859,  .859,  .859}32.9 \\
    SCORER+CBR & \cellcolor[rgb]{ .788,  .788,  .788}\textbf{56.3} & \cellcolor[rgb]{ .788,  .788,  .788}\textbf{41.2} & \cellcolor[rgb]{ .788,  .788,  .788}\textbf{74.5} & \cellcolor[rgb]{ .788,  .788,  .788}\textbf{126.8} & \cellcolor[rgb]{ .788,  .788,  .788}\textbf{33.3} \\
    \bottomrule
    \end{tabular}%
    \end{center}
    \caption{Ablation on CLEVR-Change under Total Performance.}
  \label{ablation}%
\end{table}%

\begin{table*}[h]
  \centering
  \begin{center}
    \begin{tabular}{c|ccccc|ccccc}
    \toprule
          & \multicolumn{5}{c|}{Semantic Change}  & \multicolumn{5}{c}{Only Pseudo Change} \\
    \midrule
    Method & B     & M     & R     & C     & S     & B     & M     & R     & C     & S \\
    \midrule
    Subtraction & \cellcolor[rgb]{ .953,  .953,  .953}50.2 & \cellcolor[rgb]{ .953,  .953,  .953}34.1 & \cellcolor[rgb]{ .953,  .953,  .953}67.1 & 108.0 & \cellcolor[rgb]{ .953,  .953,  .953}28 & \cellcolor[rgb]{ .953,  .953,  .953}57.3 & \cellcolor[rgb]{ .953,  .953,  .953}48.4 & \cellcolor[rgb]{ .953,  .953,  .953}74.7 & \cellcolor[rgb]{ .953,  .953,  .953}113.8 & \cellcolor[rgb]{ .929,  .929,  .929}34.0 \\
    RR    & \cellcolor[rgb]{ .929,  .929,  .929}53.3 & \cellcolor[rgb]{ .929,  .929,  .929}37.1 & \cellcolor[rgb]{ .929,  .929,  .929}70.8 & \cellcolor[rgb]{ .929,  .929,  .929}119.1 & \cellcolor[rgb]{ .929,  .929,  .929}30.4 & \cellcolor[rgb]{ .859,  .859,  .859}61.1 & \cellcolor[rgb]{ .859,  .859,  .859}50.7 & \cellcolor[rgb]{ .859,  .859,  .859}76.4 & \cellcolor[rgb]{ .929,  .929,  .929}114.9 & \cellcolor[rgb]{ .859,  .859,  .859}34.6 \\
    SCORER & \cellcolor[rgb]{ .859,  .859,  .859}54.3 & \cellcolor[rgb]{ .859,  .859,  .859}37.5 & \cellcolor[rgb]{ .859,  .859,  .859}71.5 & \cellcolor[rgb]{ .859,  .859,  .859}122.0 & \cellcolor[rgb]{ .859,  .859,  .859}31.2 & \cellcolor[rgb]{ .859,  .859,  .859}61.4 & \cellcolor[rgb]{ .859,  .859,  .859}50.6 & \cellcolor[rgb]{ .859,  .859,  .859}76.5 & \cellcolor[rgb]{ .859,  .859,  .859}116.4 & \cellcolor[rgb]{ .859,  .859,  .859}34.7 \\
    RR+CBR & \cellcolor[rgb]{ .859,  .859,  .859}54.1 & \cellcolor[rgb]{ .859,  .859,  .859}37.4 & \cellcolor[rgb]{ .859,  .859,  .859}71.5 & \cellcolor[rgb]{ .788,  .788,  .788}\textbf{122.4} & \cellcolor[rgb]{ .859,  .859,  .859}31.2 & \cellcolor[rgb]{ .929,  .929,  .929}60.7 & \cellcolor[rgb]{ .859,  .859,  .859}51.2 & \cellcolor[rgb]{ .859,  .859,  .859}76.9 & \cellcolor[rgb]{ .929,  .929,  .929}114.9 & \cellcolor[rgb]{ .859,  .859,  .859}34.6 \\
    SCORER+CBR & \cellcolor[rgb]{ .788,  .788,  .788}\textbf{54.4} & \cellcolor[rgb]{ .788,  .788,  .788}\textbf{37.6} & \cellcolor[rgb]{ .788,  .788,  .788}\textbf{71.7} & \cellcolor[rgb]{ .788,  .788,  .788}\textbf{122.4} & \cellcolor[rgb]{ .788,  .788,  .788}\textbf{31.6} & \cellcolor[rgb]{ .788,  .788,  .788}\textbf{62.0} & \cellcolor[rgb]{ .788,  .788,  .788}\textbf{51.7} & \cellcolor[rgb]{ .788,  .788,  .788}\textbf{77.4} & \cellcolor[rgb]{ .788,  .788,  .788}\textbf{117.9} & \cellcolor[rgb]{ .788,  .788,  .788}\textbf{35.0} \\
    \bottomrule
    \end{tabular}%
    \end{center}
    \caption{Ablation study on CLEVR-Change under the evaluation of semantic change and only pseudo change.}
  \label{change}%
\end{table*}%

\subsection{Ablation Study and Analysis}
\textbf{Ablation Study of Each Module on CLEVR-Change.} Table \ref{ablation} shows ablation study of each module under total performance.  Subtraction indicates directly subtracting two images; RR means vanilla representation reconstruction. We find that RR is much better than Subtraction, showing that match-based strategy is more reliable than direct subtraction under pseudo changes. When we maximize cross-view contrastive alignment of two images, SCORER yields a further performance boost. This shows that it is important to learn the representations invariant under pseudo changes, which is key to learn a stable difference representation.  Besides, when we augment RR and SCORER with CBR, both RR+CBR and SCORER+CBR achieve better performances. This not only validates that CBR improves captioning quality, but also proves that CBR is generalizable. 

Table \ref{change} shows the ablation study of each module under semantic change and only pseudo change, separately. We can obtain observations similar to the total performance.  Besides, we find that SCORER is much better than RR under semantic change, but under only pseudo change, SCORER
brings less gain. This results from that in this case, the learned difference representation contains less information, making SCORER difficult to align it with words. By contrast, SCORER+CBR significantly improves RR on both settings, which shows that SCORER and CBR supplement each other. 
More ablation studies on the other datasets are in the supplementary material.

\begin{figure}[t]
  \centering
  \includegraphics[width=1\linewidth]{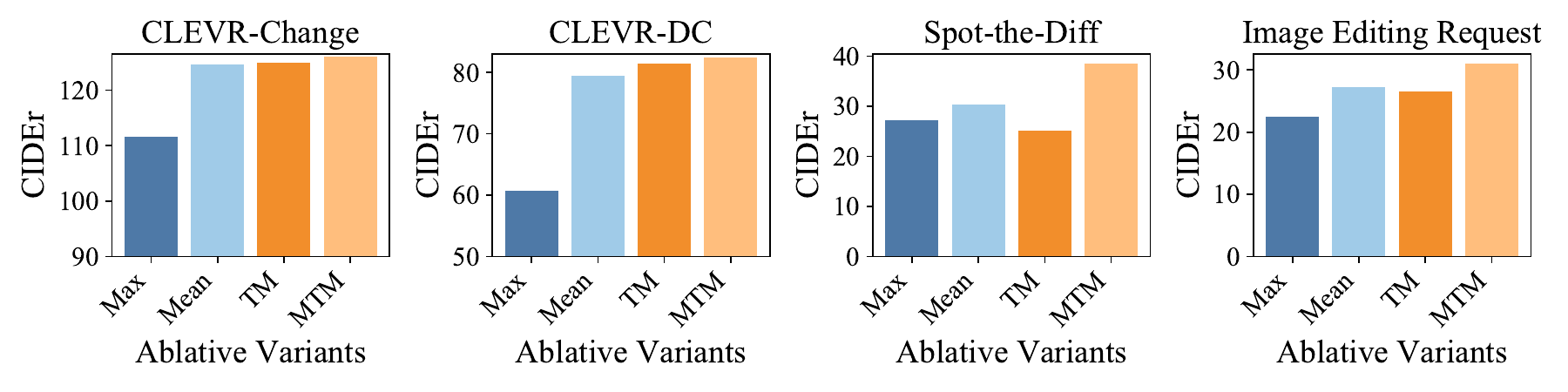}
   \caption{Ablation studies of MTM on four datasets.}
   \label{MTM}
\end{figure}

\begin{figure}[t]
  \centering
  \includegraphics[width=1\linewidth]{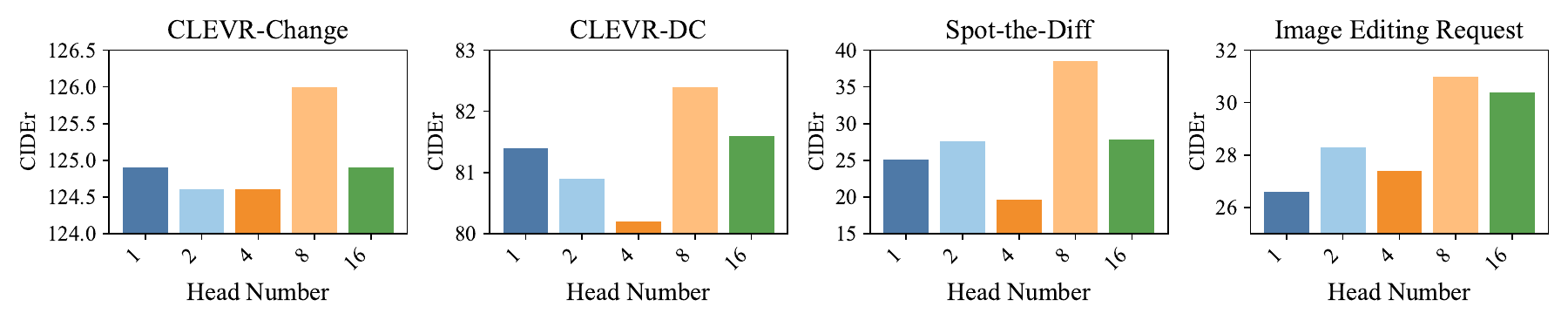}
   \caption{Effect of head number of SCORER on four datasets.}
   \label{MTM_head}
\end{figure}

\begin{figure}[t]
  \centering
  \includegraphics[width=1\linewidth]{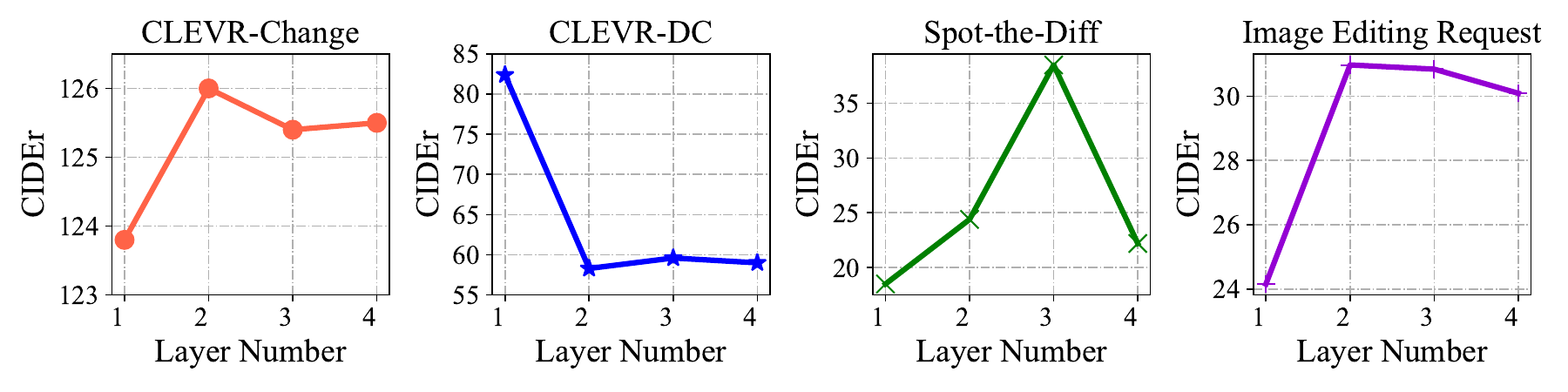}
   \caption{Effect of layer number of SCORER on four datasets.}
   \label{layer_num}
\end{figure}

\begin{figure}[h]
  \centering
  
  \includegraphics[width=1\linewidth]{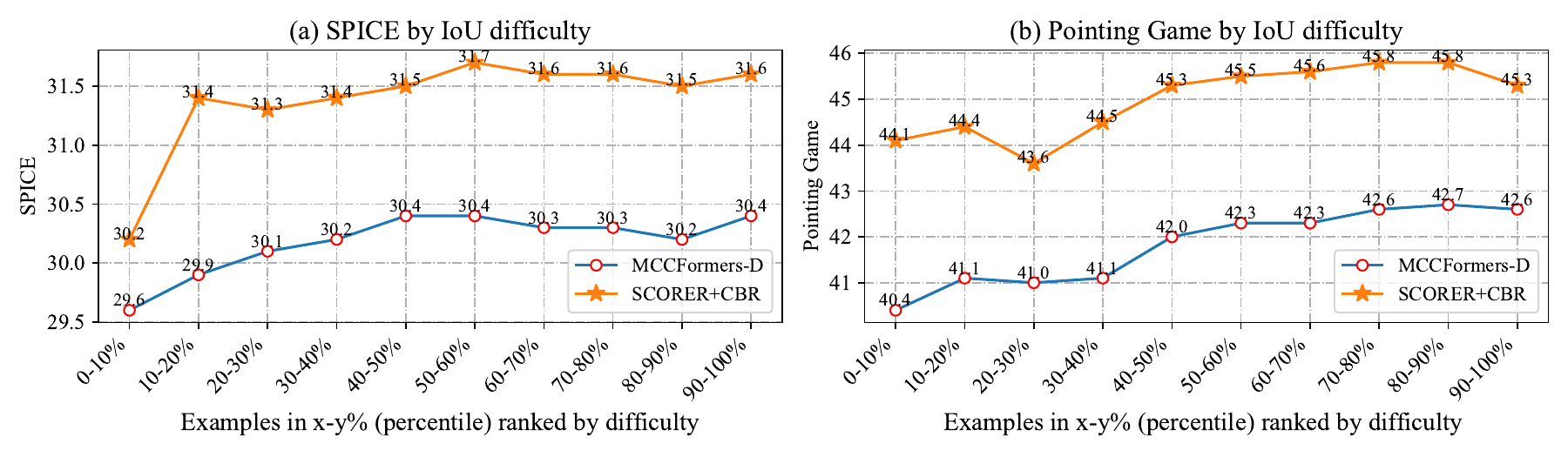}
   \caption{Captioning and change localization of varied viewpoints.}
   \label{viewpoint}
\end{figure}

\begin{figure*}[t]
\centering
\includegraphics[width=1\textwidth]{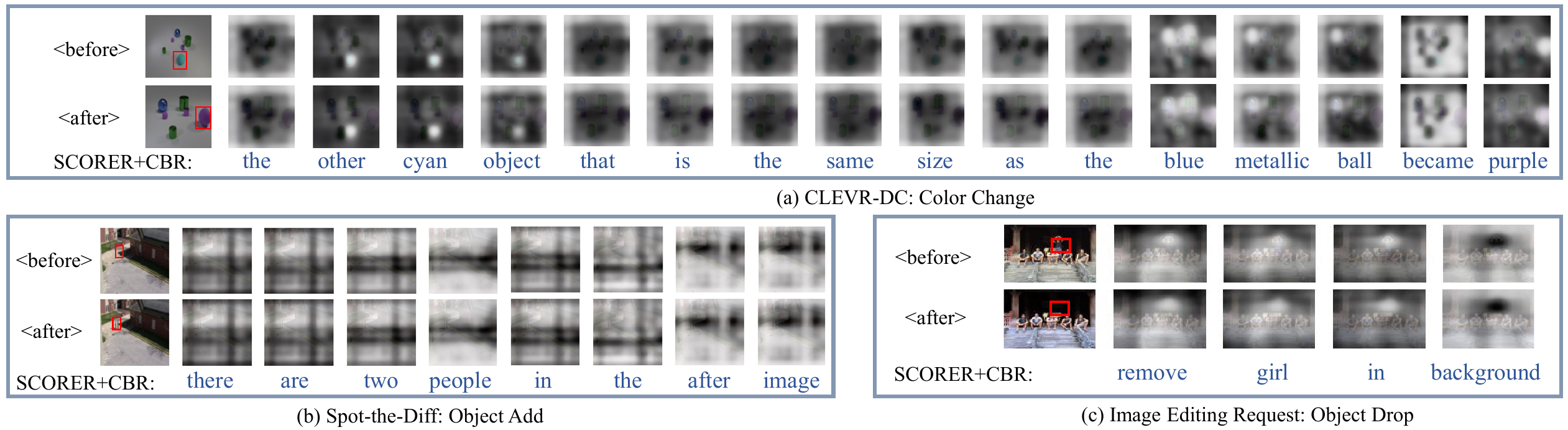} 
\caption{Three cases  from different scenarios, where the generated captions along with the attention weight at each word are visualized. }
\label{example}
\end{figure*}

\textbf{Ablation Study of MTM.} Instead of using MTM to perform fine-grained matching between different feature subspaces of cross-view images, we use max/mean-pooling to obtain the global feature of each image and compute their similarity. Besides, we implement TM without multi-head operation. The results in Fig. \ref{MTM} show that MTM achieves the best results, which demonstrates that it plays a critical role in view-invariant representation learning. Besides, only implementing token-wise matching is not better than simple mean-pooling. Our conjecture is that the changed object commonly appears in a local region with weak feature, so it is insufficient to reveal this slight difference by only interacting features at token level. As such, it is necessary to match two image features at finer level, \emph{i.e.,} subspace level. 


\textbf{Effect of Head Number of SCORER.} We further investigate the effect of head number for SCORER, \textit{i.e.}, the head number of MTM and MHCA (Eq. (\ref{mhca})). The results are shown in Fig. \ref{MTM_head}. We find that the best results are achieved on the four datasets when setting the head number as 8.

\textbf{Effect of Layer Number of SCORER.} We investigate the effect of layer number for SCORER in Fig. \ref{layer_num}. On four datasets, we find that increasing the layer number does not bring better performance, because deeper layers could result in the problem of over-fitting. Besides, the layer number is the deepest on Spot-the-Diff. Our conjecture is that objects have no good postures and background information is more complex in a surveillance scenario. As such, we empirically set proper layer number of 2, 1, 3, and 2 on four datasets.

\begin{figure}[t]
\setlength{\belowcaptionskip}{-0.4cm}
\includegraphics[width=0.48\textwidth]{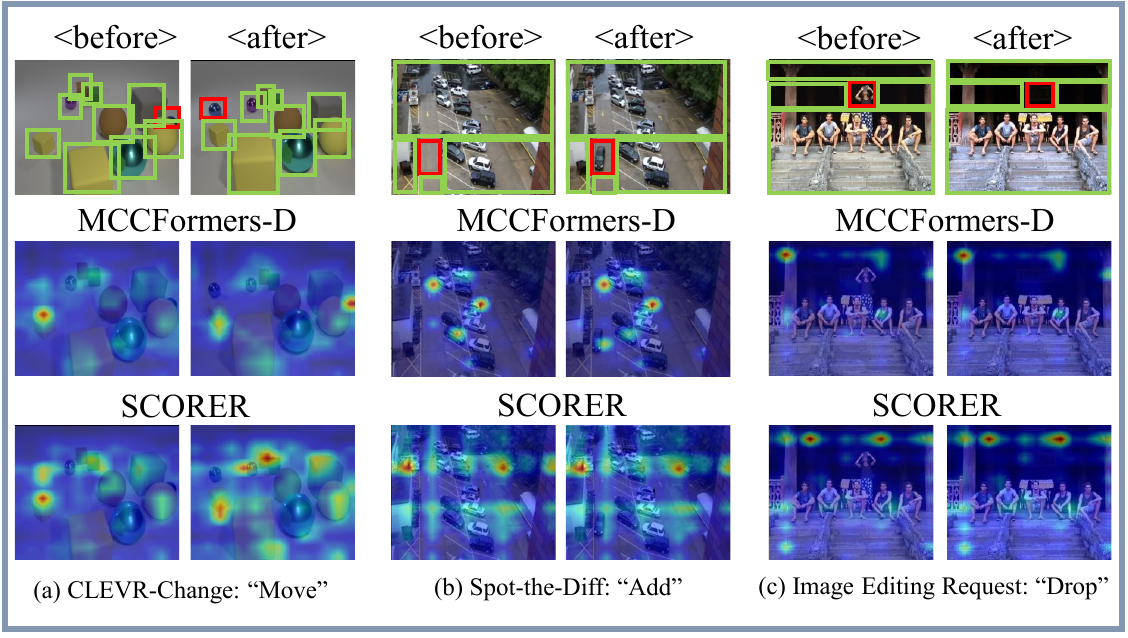} 
\caption{Visualization of the alignment of unchanged objects computed by MCCFormers-D \cite{Qiu_2021_ICCV} and our SCORER.}

\label{align}
\end{figure}

\subsection{Captioning and change localization results with varied viewpoints}
To intuitively evaluate the efficacy of  our method  to handle pseudo changes,  we show the captioning (Fig. \ref{viewpoint} (a)) and change localization (Fig. \ref{viewpoint} (b)) results of SCORER+CBR and SOTA method MCCFormers-D \cite{Qiu_2021_ICCV} with varied viewpoints. The amount of viewpoint change is measured by the IoUs of objects' bounding boxes across an image pair (lower IoU means higher difficulty).  For change localization, the pioneer work DUDA \cite{park2019robust} tried Pointing Game to evaluate attention maps of change localization, where maps are computed by using the captured difference to directly query related regions on each image. In contrast, we consider simultaneously evaluating change localization and cross-modal alignment, \emph{i.e.}, attention maps of cross-modal alignment, to check whether the model can locate changed regions when generating corresponding words. This is more challenging but more reasonable.  In Fig. \ref{viewpoint}, we find that our method outperforms MCCFormers-D and shows better robustness with varied viewpoint changes on both evaluations, which benefit from view-invariant representation learning and cross-modal backward reasoning.

\subsection{Qualitative Analysis}
To intuitively evaluate our method, we conduct qualitative analysis on the four datasets.  Fig. \ref{example} illustrates three cases in different change scenarios. For each case, we visualize the generated caption along with the attention weight at each word. When the weight is higher, the region is brighter. We observe that when generating the words about the changed object or its referents, SCORER+CBR can adaptively locate the corresponding regions.
In Fig. \ref{align}, we visualize  the alignment between unchanged objects under different change scenes. The compared method is the SOTA method MCCFormers-D \cite{Qiu_2021_ICCV}. We implement it based on the released code. We find that when directly correlating two image features, MCCFormers-D only aligns salient objects between two images. Instead, our SCORER first learns two view-invariant  representations in a self-supervised way. Based on these, SCORER can  better align and reconstruct the   representations of unchanged objects, so as to facilitate subsequent difference representation learning.  More qualitative examples  are shown in the supplementary material.

\section{Conclusion}
\label{conclusion}
This paper  proposes a novel SCORER to learn a stable difference representation while resisting pseudo changes. SCORER first learns two view-invariant image representations in a self-supervised way, by maximizing the cross-view contrastive alignment of two images.  Based on these, SCORER mines their common features to reconstruct the representations of unchanged objects. This helps learn a stable difference representation for caption generation.  Further, we design the CBR to improve captioning quality by enforcing the yielded caption is informative about the difference in a self-supervised manner. 
Extensive experiments show that our method achieves  the state-of-the-art results on four public datasets with different change scenarios.


\section*{Acknowledgements}
This work was supported by the National Key Research and Development Program of China under Grant (2018AAA0102000), National Nature Science Foundation of China (62322211, U21B2024, 61931008, 62071415, 62236008, U21B2038), Fundamental Research Funds for the Central Universities, ``Pioneer'', Zhejiang Provincial Natural Science Foundation of China (LDT23F01011F01, LDT23F01015F01, LDT23F01014F01) and ``Leading Goose'' R\&D Program of Zhejiang Province (2022C01068), and  Youth Innovation Promotion Association of Chinese Academy of Sciences (2020108). 
{\small
\bibliographystyle{ieee_fullname}
\bibliography{egbib}
}

\section{Supplementary Material}
\label{sec:intro}
In this supplementary material, we will show more experimental results. First, we show the implementation details on the four datasets. Second, we provide the discussion of trade-off parameters on the four datasets. Next, we show more ablation studies on the four datasets.  Finally, we show more qualitative examples on the four datasets. 

\subsection{Implementation Details}
We provide more implementation details of our method. During training, the batch sizes and learning rates of our method on the four datasets are shown in Table \ref{batch}.  We train the model to convergence with 10K iterations in total. Both training and inference are implemented with PyTorch on an RTX 3090 GPU. The used resources on the four datasets are shown in Table \ref{training time}. We can find that our method does not need much training time and GPU memory, so it can be easily reproduced by   other researchers.
\begin{table}[h]
			\centering
			\begin{center}
			    
    \begin{tabular}{c|c|c}
   \hline
          & Batch Size & Learning Rate \\
    \hline
    CLEVR-Change & 128   &  2 $\times$ $10^{-4}$ \\
    CLEVR-DC & 128  & 2 $\times$ $10^{-4}$ \\
    Spot-the-Diff & 32 &   2 $\times$ $10^{-4}$ \\
    Image Editing Request & 16 &  1 $\times$ $10^{-4}$  \\
    \hline
    \end{tabular}%
    \end{center}
    \caption{The training parameters on the four datasets.}
\label{batch}%
		
	\end{table}

\begin{table}[htbp]
  \centering
   \begin{center}
  
    \begin{tabular}{c|c|c}
   \hline
    \multicolumn{1}{c|}{} & Training Time & GPU Memory \\
    \hline
    CLEVR-Change & 3 hours & 20G \\
    CLEVR-DC & 1.5 hours & 15.6G \\
    Spot-the-Diff & 20 minutes & 5G \\
    Image Editing Request & 15 minutes  & 4.3G \\
 \hline
    \end{tabular}%
    \end{center}
    \caption{Used training time and GPU memory on the four datasets.}
  \label{training time}%
\end{table}%

\begin{figure}[htbp]

\centering
\includegraphics[width=0.49\textwidth]{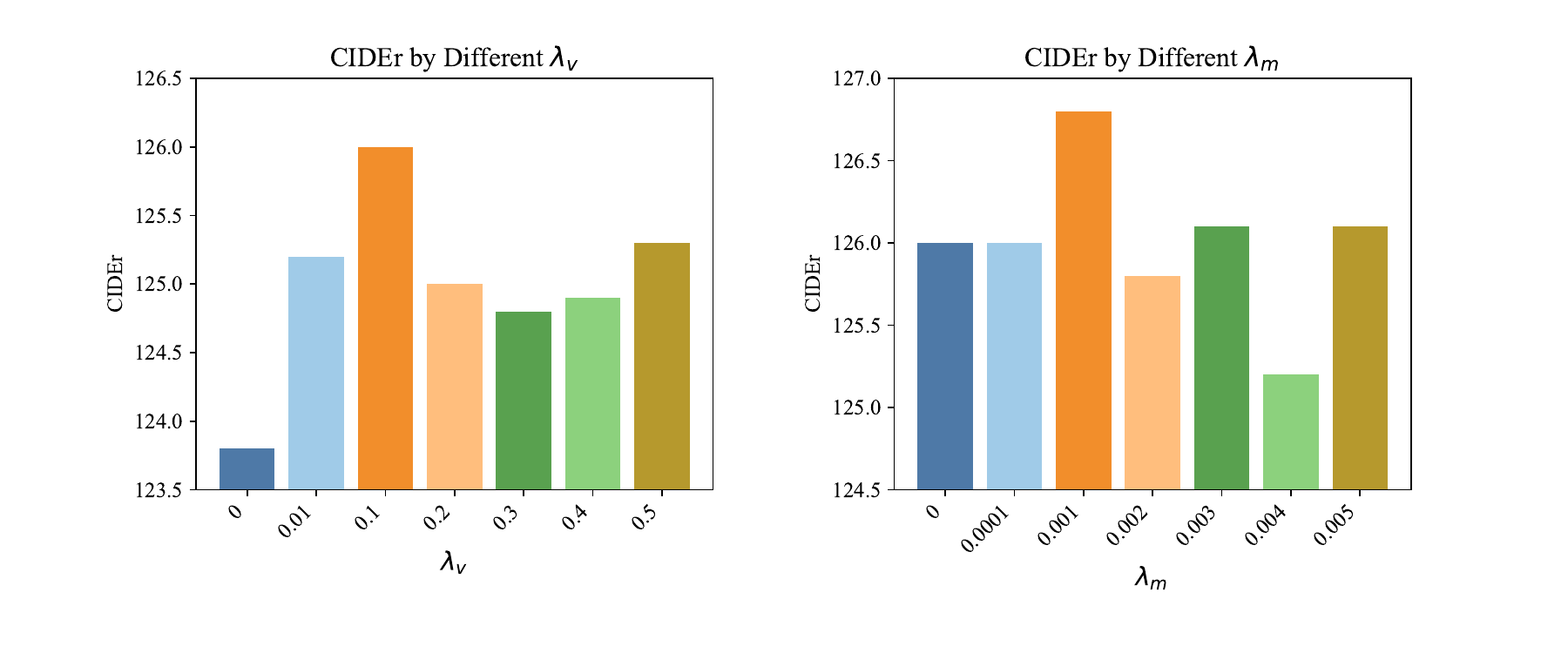} 
\caption{The effects of $\lambda_v$ and $\lambda_m$ on CLEVR-Change. }
\label{cider_change}
\end{figure}

\begin{figure}[htbp]

\centering
\includegraphics[width=0.49\textwidth]{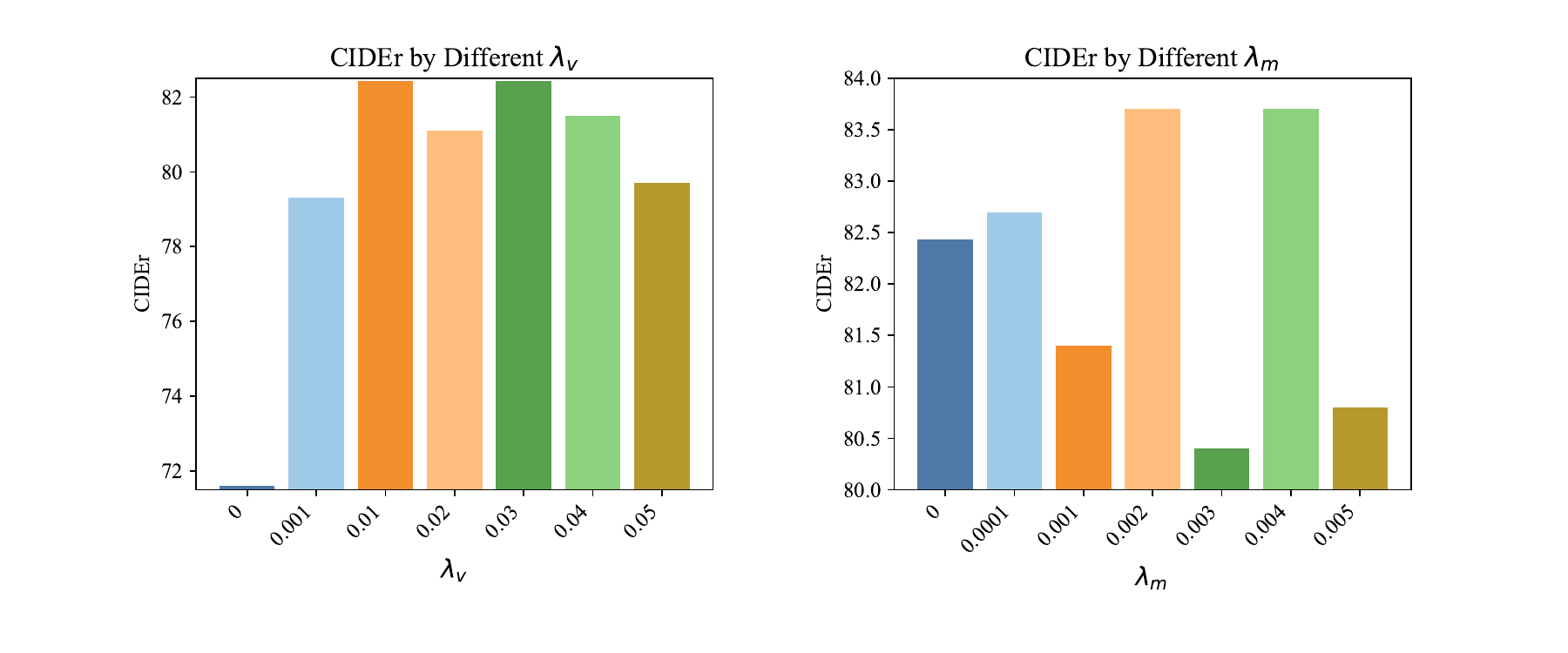} 
\caption{The effects of $\lambda_v$ and $\lambda_m$ on CLEVR-DC. }
\label{cider_dc}
\end{figure}

\begin{figure}[htbp]

\centering
\includegraphics[width=0.49\textwidth]{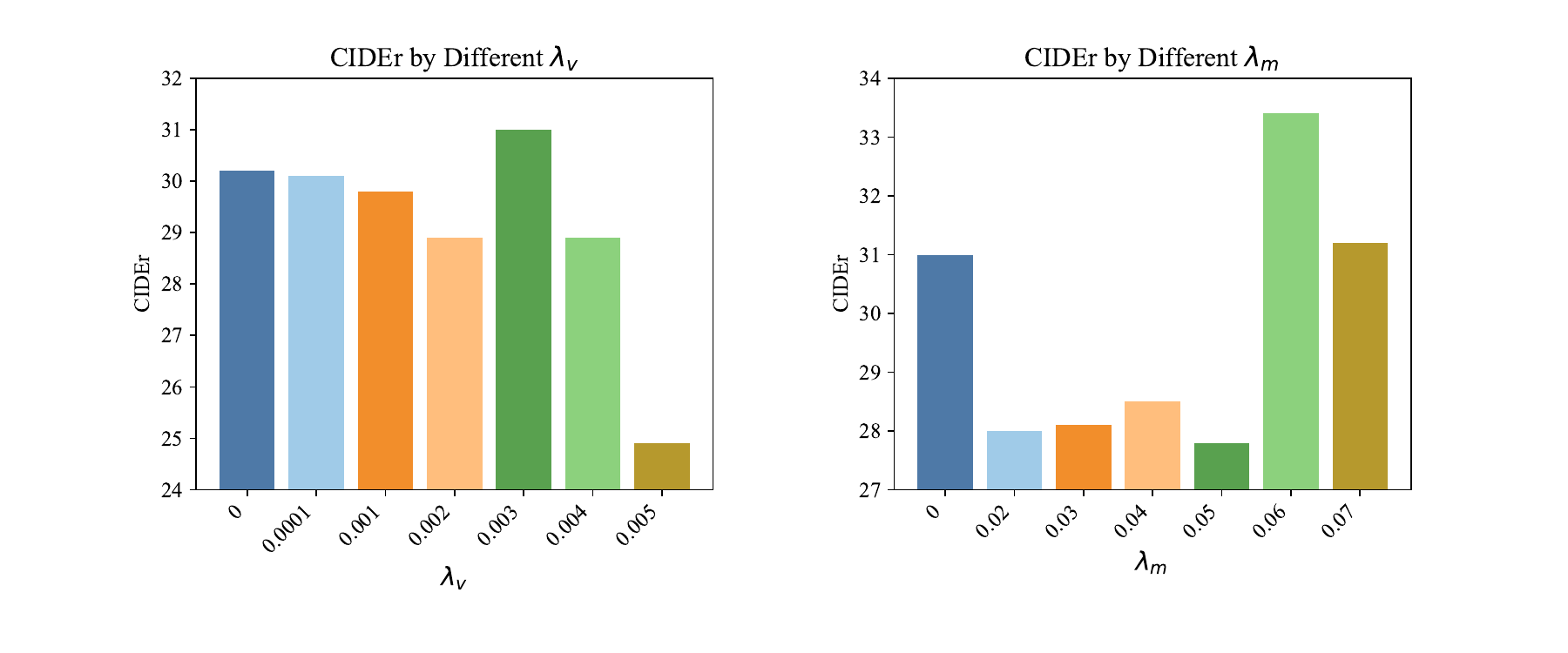} 
\caption{The effects of $\lambda_v$ and $\lambda_m$ on Image Editing Request. }
\label{cider_spot}
\end{figure}

\begin{figure}[htbp]

\centering
\includegraphics[width=0.49\textwidth]{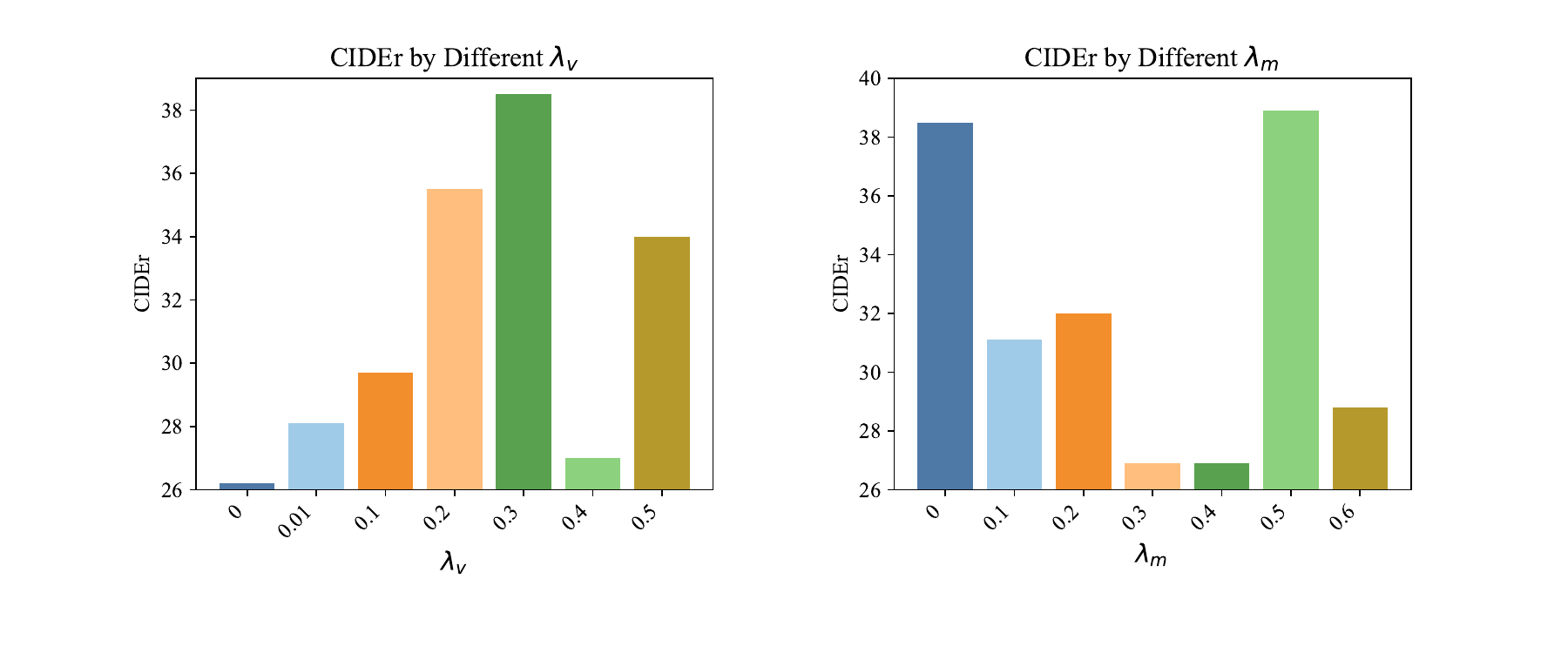} 
\caption{The effects of $\lambda_v$ and $\lambda_m$ on Spot-the-Diff. }
\label{cider_spot}
\end{figure}

\begin{table}[t]
\begin{center}
  \centering
 
    \begin{tabular}{c|c|c|c|c|c}
    \toprule
    Ablation & B     & M     & R     & C     & S \\ \midrule
    Subtraction & 46.2  & 31.5  & 63.4  & 68.5  & 13.9 \\
    RR    & 46.9  & 31.7  & 64.2  & 71.6  & 14.6 \\
    SCORER & \textbf{49.5}  & \textbf{33.4}  & 66.0  & 82.4  & 15.8 \\
    RR+CBR & 47.2  & 32.3  & 64.4  & 73.0  & 15.0 \\
    SCORER+CBR & 49.4 & \textbf{33.4} & \textbf{66.1} & \textbf{83.7} & \textbf{16.2} \\
    \bottomrule
    \end{tabular}%
    \end{center}
     \caption{Ablation study on CLEVR-DC.}
  \label{dc}%
  \end{table}

  \begin{table}[t]
\begin{center}
  \centering
 
    \begin{tabular}{c|c|c|c|c|c}
    \toprule
    Ablation & B     & M     & R     & C     & S \\
    \midrule
    Subtraction & 6.7   & 13.7  & 37.4  & 22.1  & 8.7 \\
    RR    & 8.3   & 14.3  & 39.2  & 30.2  & 12.4 \\
    SCORER & 9.6   & 14.6  & 39.5  & 31.0  & \textbf{12.6} \\
    RR+CBR & 7.1   & 14.6  & 40.6  & 31.9  & 12.3 \\
    SCORER+CBR & \textbf{10.0} & \textbf{15.0} & \textbf{39.6} & \textbf{33.4} & \textbf{12.6} \\
    \bottomrule
    \end{tabular}%
    \end{center}
     \caption{Ablation study on Image Editing Request.}
  \label{edit}%
  \end{table}

\begin{table}[t]
\begin{center}
  \centering
    \begin{tabular}{c|c|c|c|c|c}
    \toprule
    Ablation & B     & M     & R     & C     & S \\
    \midrule
    Subtraction & 7.2   & 11.9  & 28.9  & 29.4  & 13.0 \\
    RR    & 7.4   & 12.0  & 27.6  & 26.2  & 14.2 \\
    SCORER & 9.4   & \textbf{13.8} & \textbf{32.0} & 38.5  & \textbf{19.3} \\
    RR+CBR & 8.1   & 11.7  & 30.6  & 32.4  & 15.6 \\
    SCORER+CBR & \textbf{10.2} & 12.2  & 31.9  & \textbf{38.9} & 18.4 \\
    \bottomrule
    \end{tabular}%
    \end{center}
    \caption{Ablation study on Spot-the-Diff.}
  \label{spot}%

\end{table}

\subsection{Study on the Trade-off Parameters}

In this section, we discuss the trade-off parameters $\lambda_v$ and $\lambda_m$ in Eq.~(13) of the main paper  on the four datasets. Both parameters are to balance the contributions from the caption generator, SCORER, and CBR. In Figure \ref{cider_change}, We first analyze the effect of $\lambda_v$ on the CLEVR-Change dataset. We find that the performance of SCORER changes under different values,  because the   model will focus much on one part but ignore the supervision from the other. We empirically set $\lambda_v$ to 0.1. Then, we fix  $\lambda_v$ to discuss the effect of $\lambda_m$ for SCORER+CBR and  set it as $0.001$. In Figure \ref{cider_dc}, we also first analyze the effect of $\lambda_v$ on the CLEVR-DC dataset. We empirically set $\lambda_v$ to 0.01. Then, we fix $\lambda_v$ to discuss $\lambda_m$. We find that the CIDEr scores are identical when setting $\lambda_m$ as 0.002 and 0.004. Thus, we further compare their SPICE scores (16.0 vs. 16.2) and set $\lambda_m$ to 0.004. By that analogy, we discuss the effect of $\lambda_v$ and $\lambda_m$ on the Image Editing Request and Spot-the-Diff datasets. We empirically set $\lambda_v$ and $\lambda_m$ as 0.003 and 0.06 on Image Editing Request; 0.3 and 0.5 on Spot-the-Diff.

\subsection{Ablation Study}
We carry out ablation studies to validate the effectiveness of our method. (1) Subtraction is a transformer-based baseline model which computes difference features by direct subtraction.
(2) RR refers to vanilla representation reconstruction without cross-view contrastive alignment. (3) SCORER is the proposed self-supervised cross-view representation reconstruction network. (4) CBR means the proposed module of cross-modal backward reasoning.


\textbf{Results on CLEVR-DC.} Table \ref{dc} shows the ablation studies of our method on the CLEVR-DC dataset, which are evaluated in terms of total performance. We can draw the
same conclusion  from the ablative variants. Compared with the baseline model of Subtraction, it is effective to  first compute the aligned properties and then deduce the difference features. The proposed SCORER first learns the representations that are invariant under extreme viewpoint changes for a pair of similar images,  by maximizing their cross-view contrastive alignment. Then, SCORER can fully mine their common features to reconstruct the representations of unchanged objects, thereby learning a stable difference representation for caption generation.
Besides, CBR is helpful to improve the quality of generated sentences, which shows that it does enforce the yielded sentence to be informative about the learned difference.

\textbf{Results on Image Editing Request.} Table \ref{edit} shows the ablation studies of our method on the Image Editing Request dataset, where two images in the pair are aligned and the edited objects on this dataset are usually inconspicuous. We can obtain the same observations. Match-based strategy (RR) performs better than the strategy of direct subtraction. The proposed SCORER can fully align and mine the common features between two images, so as to reconstruct reliable unchanged representations for learning a stable difference representation. When we implement CBR, the performance of SCORER+CBR is further boosted, which shows that CBR is helpful to improve captioning quality.

\textbf{Results on Spot-the-Diff.} Table \ref{spot} shows the ablation studies of our method on the Spot-the-Diff dataset. We can find that compared with the baseline model of Subtraction, the improvement is not  significant when using the model of representation reconstruction. Our conjecture is that image pairs on this dataset are well-aligned, so direct subtraction also can capture some salient changes. When we perform cross-view contrastive alignment, the performance of SCORER is significantly boosted, which shows that it facilitates correct alignment between unchanged objects, so as to help locate fine-grained changes. When introducing CBR, we observe that the results of RR+CBR and SCORER+CBR are not improved  significantly. As we discussed in the main paper, the image pairs on this dataset actually contain one or more changes. For fair-comparison, we conduct experiments mainly based on the single-change setup. This makes  the ``hallucination'' representation, which is reversely modeled by the ``before'' representation and single-change caption, not fully matched with the ``after'' representation. In this situation, the performances of RR+CBR and SCORER+CBR do not gain significant improvement. 

\begin{figure}[t]
  \centering
  \includegraphics[width=1\linewidth]{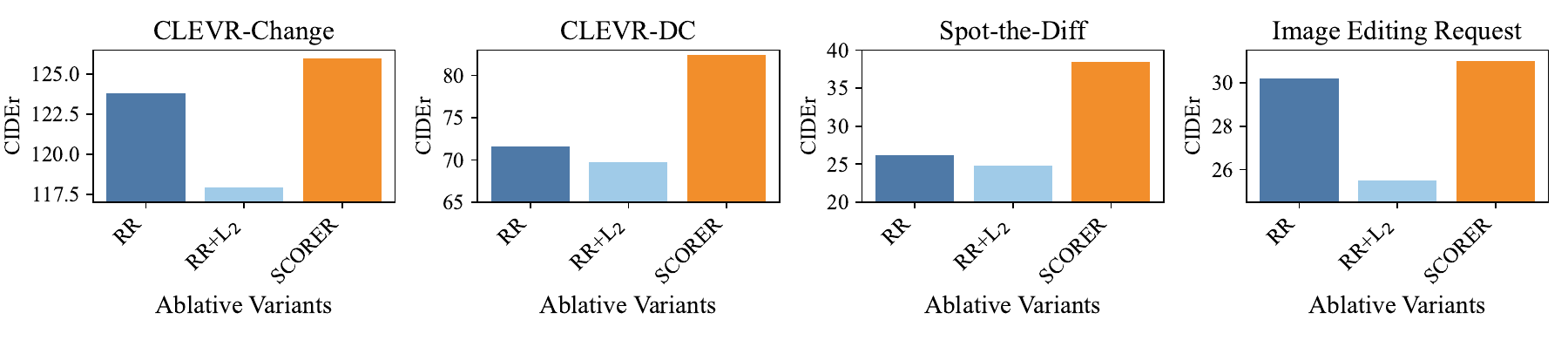}
   \caption{Effect of cross-view contrastive alignment on four datasets.}
   \label{l2}
\end{figure}

\textbf{Effect of Cross-view Contrastive Alignment.} We study the effect of cross-view contrastive alignment, which is key to  learn view-invariant image representations. Besides, we try $L_2$ distance metric to achieve this goal by only maximizing alignment of similar images. Fig. \ref{l2} shows comparison results among RR (without alignment constraint), RR+$L_2$, and SCORER. We find that SCORER achieves the best result, while the performance of RR+$L_2$ is the worst.  The comparison results validate that it is necessary to build contrastive alignment between similar/dissimilar images, which helps the model focus more on the change of feature and resist feature shift. As a result, the model can capture the stable difference representation between two images for caption generation. 

\textbf{Study of Different Fusion Strategies to Model Difference Representation.}  In the Eq. (7) of main paper, we obtain the difference representation between two images by concatenating the changed features of each image. In order to validate whether concatenation is a good choice in the case of extreme viewpoint changes, besides concatenation (``cat''), we  try to use other fusion strategies to model the difference representation between two images: sum,  hadamard product, and respective interaction with words and then ``cat''. The experiment is conducted on the CLEVR-DC dataset with extreme viewpoint changes. Here, we report the comparison results under CIDEr metric:  ``cat'' (82.4),  sum (61.9), hadamard product (61.6), respective interaction with words and then ``cat'' (81.3). The results validate the effectiveness of our choice:  using ``cat'' to construct an omni-representation of change between cross-view images. With this omni-representation, our model can accurately locate changed regions on two images during word generation, even under extreme viewpoint changes. This benefits from view-invariant representation learning and cross-modal backward reasoning.

\subsection{Qualitative Analysis}
In this section, we provide more visualization results about the alignment of unchanged objects, and the generated captions along with the attention weight at each word on the four datasets.

\begin{figure*}[htbp]
\centering
\includegraphics[width=1\textwidth]{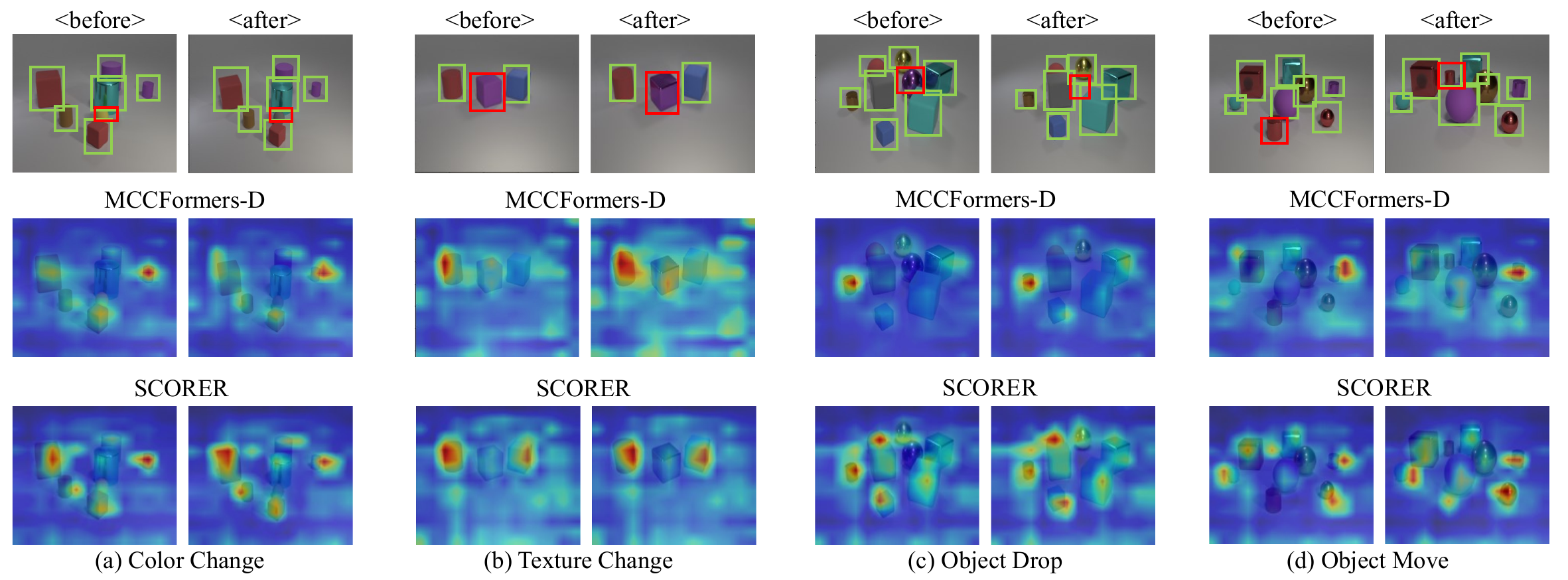} 
\caption{Visualization of the alignment of unchanged objects on CLEVR-Change,  computed by MCCFormers-D and our SCORER. }
\label{align_change}
\end{figure*}

\begin{figure*}[htbp]
\centering
\includegraphics[width=1\textwidth]{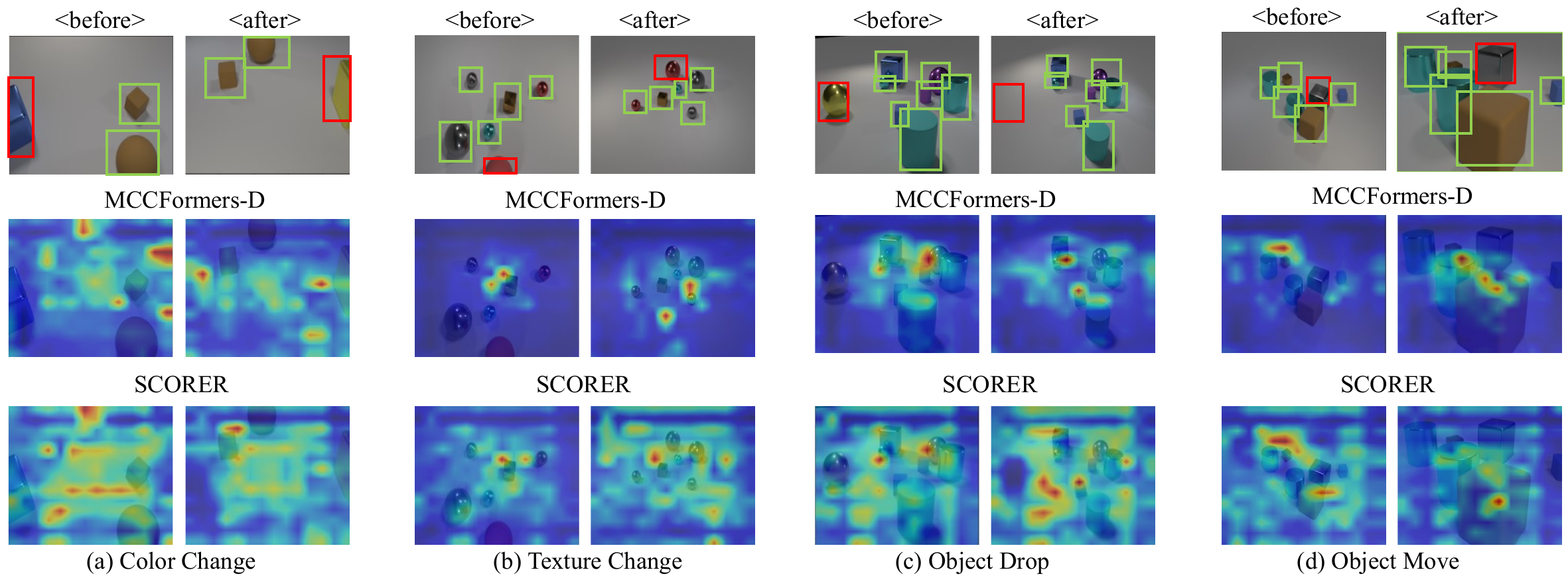} 
\caption{Visualization of the alignment of unchanged objects on CLEVR-DC, computed by MCCFormers-D and our SCORER. }
\label{align_dc}
\end{figure*}

\begin{figure*}[htbp]
\centering
\includegraphics[width=1\textwidth]{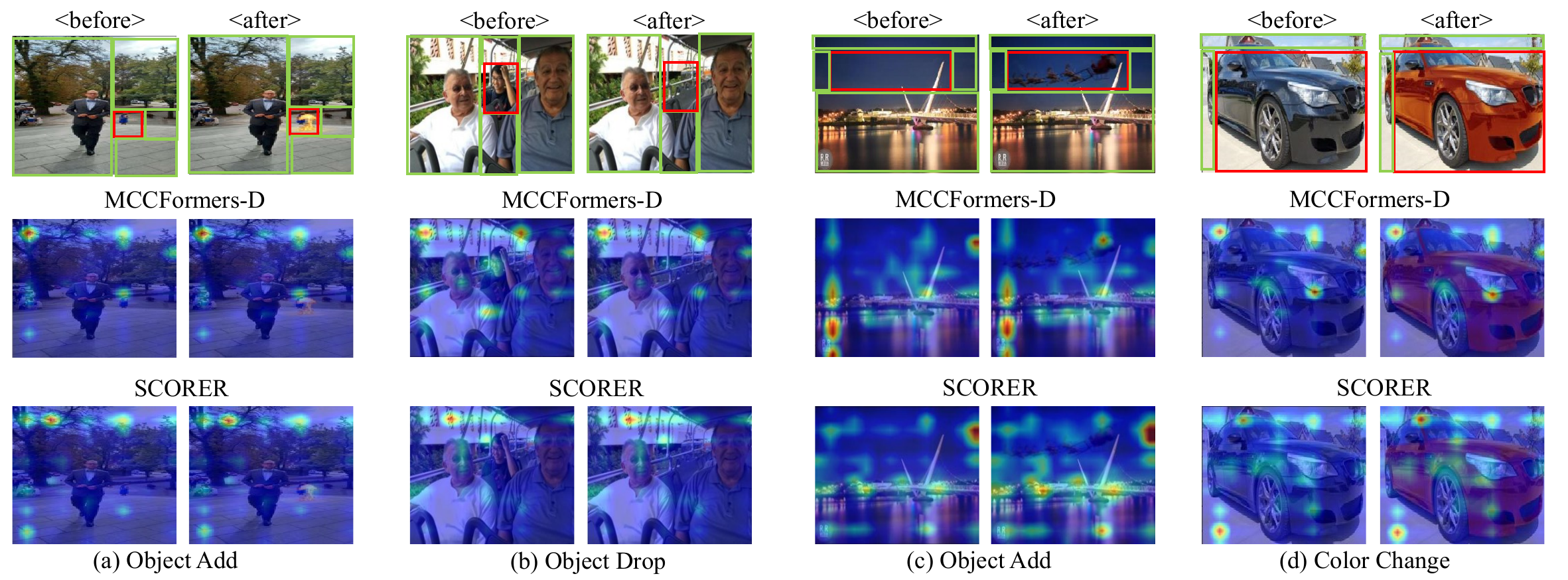} 
\caption{Visualization of the alignment of unchanged objects on Image Editing Request, computed by MCCFormers-D and our SCORER. }
\label{align_edit}
\end{figure*}

\begin{figure*}[htbp]
\centering
\includegraphics[width=1\textwidth]{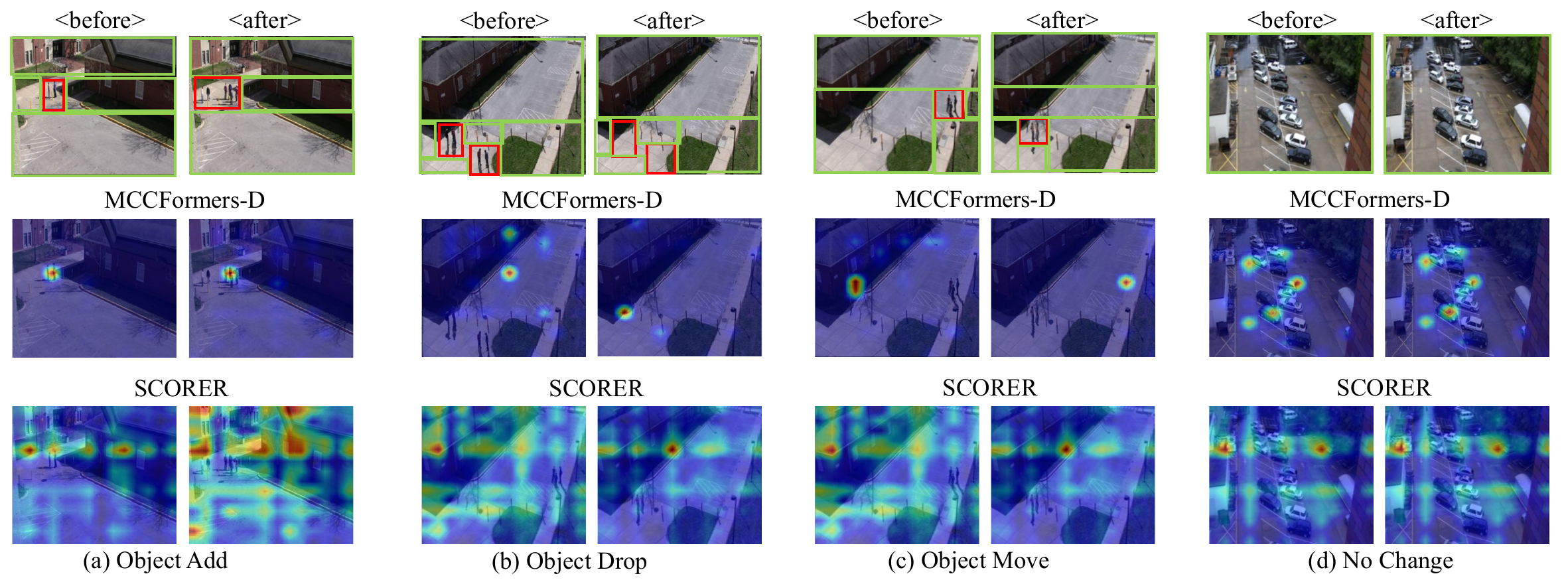} 
\caption{Visualization of the alignment of unchanged objects on Spot-the-Diff, computed by MCCFormers-D and our SCORER.  }
\label{align_spot}
\end{figure*}

In Figure \ref{align_change} - \ref{align_spot}, we visualize  the alignment between unchanged objects under different change types on CLEVR-Change, CLEVR-DC, Image Editing Request, and Spot-the-Diff, respectively. The compared method is MCCFormers-D that is a state-of-the-art method based on transformer. To fully match cross-view images,  both SCORER and MCCFormers-D respectively use one image to query the shared objects on the other one, so obtaining two attention maps about cross-view alignment. We find that when directly matching two image features, MCCFormers-D mainly attends to some salient objects. Instead, our SCORER first learns two view-invariant image representations in a self-supervised way, by maximizing their cross-view contrastive alignment.  Based on these, SCORER can better align and reconstruct the representations of unchanged objects, so as to facilitate subsequent difference representation learning.

\begin{figure*}[htbp]
\centering
\includegraphics[width=1\textwidth]{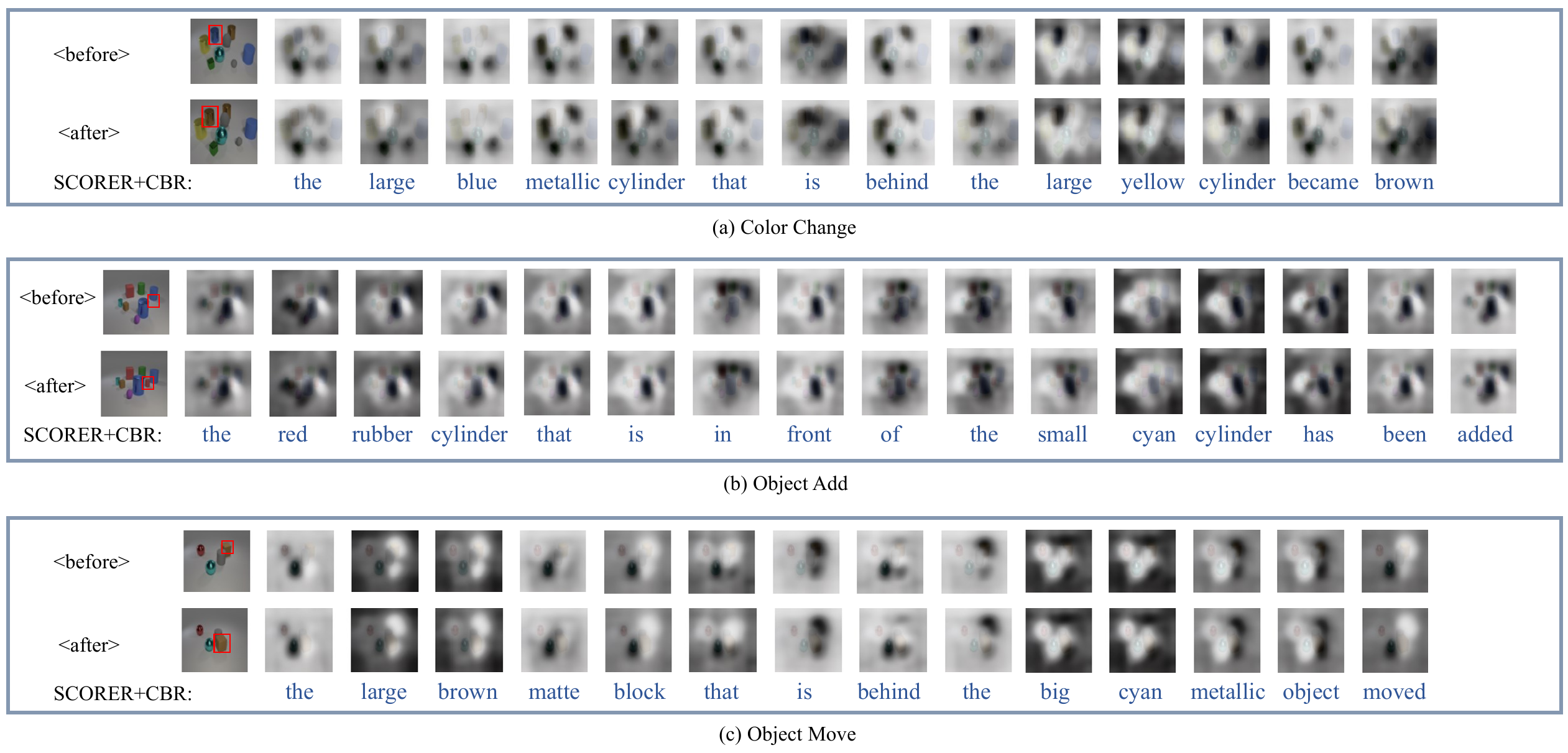} 
\caption{Three cases about ``Color Change'', ``Add'', and ``Move'' from CLEVR-Change, where the generated captions along with the attention weight  at each word are visualized. }
\label{word_change}
\end{figure*}

In Figure \ref{word_change} - \ref{word_spot}, we visualize the generated captions along with the attention weight at each word under different change types on CLEVR-Change, CLEVR-DC, Image Editing Request, and Spot-the-Diff, respectively. When predicting the next word, the decoder uses generated words to compute attention over the learned difference representation, which yields a single attention map about cross-modal alignment. We interpolate it on each image to show the localization of before- and after-changed object during  word generation. When the attention weight is higher, the localized region is brighter. We observe that when generating the words about the changed object or its referent, SCORER+CBR can adaptively attend to the corresponding regions. This superiority mainly results from the facts that 1) SCORER learns two view-invariant image representations for reconstructing the representations of unchanged objects, so as to learn a stable difference representation for caption generation; 2) cross-modal backward reasoning can improve the quality of generated captions by enforcing the caption to be informative about the learned difference.

\begin{figure*}[htbp]
\centering
\includegraphics[width=1\textwidth]{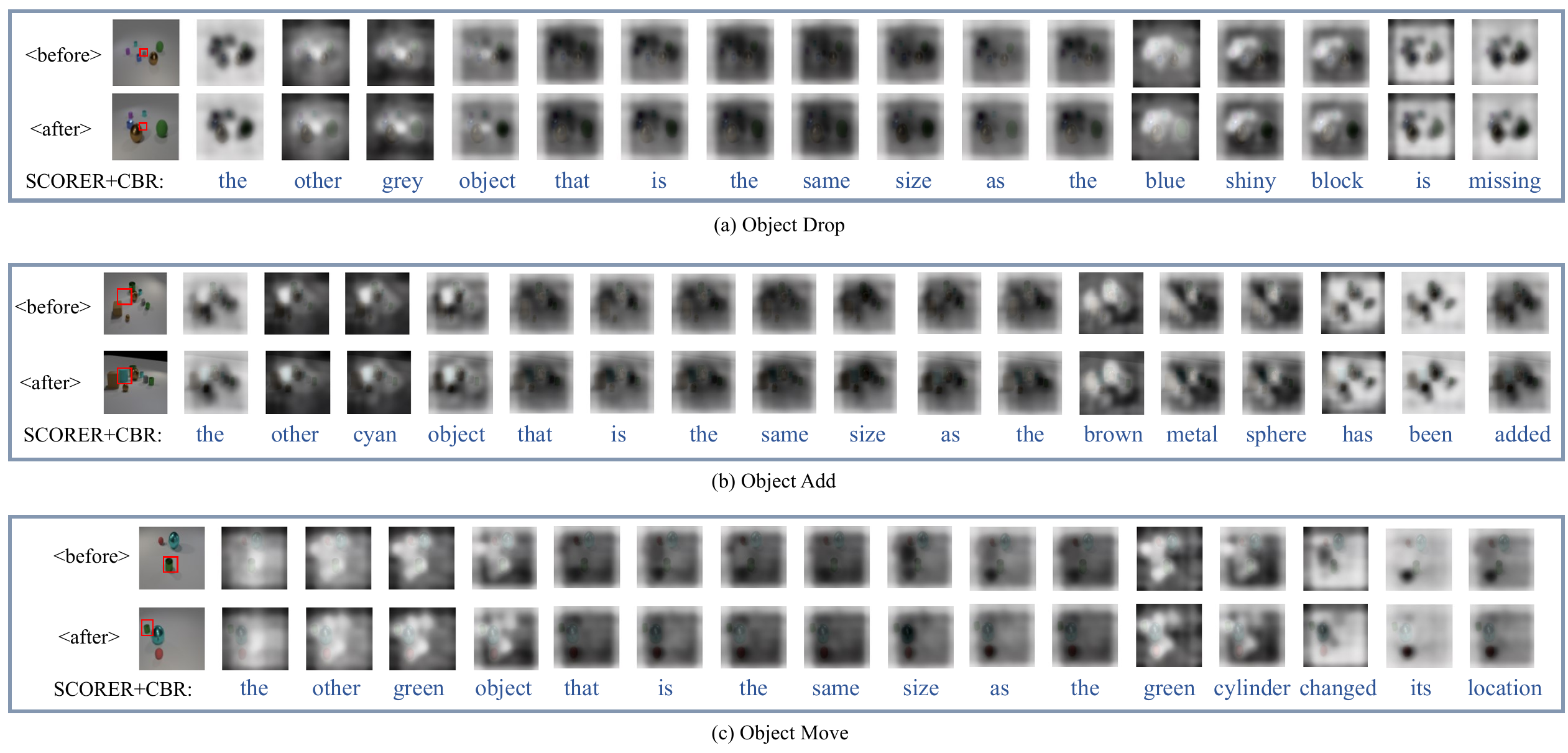} 
\caption{Three cases about ``Drop'', ``Add'', and ``Move'' from CLEVR-DC, where the generated captions along with the attention weight  at each word are visualized.  }
\label{word_dc}
\end{figure*}

\begin{figure*}[htbp]
\centering
\includegraphics[width=0.7\textwidth]{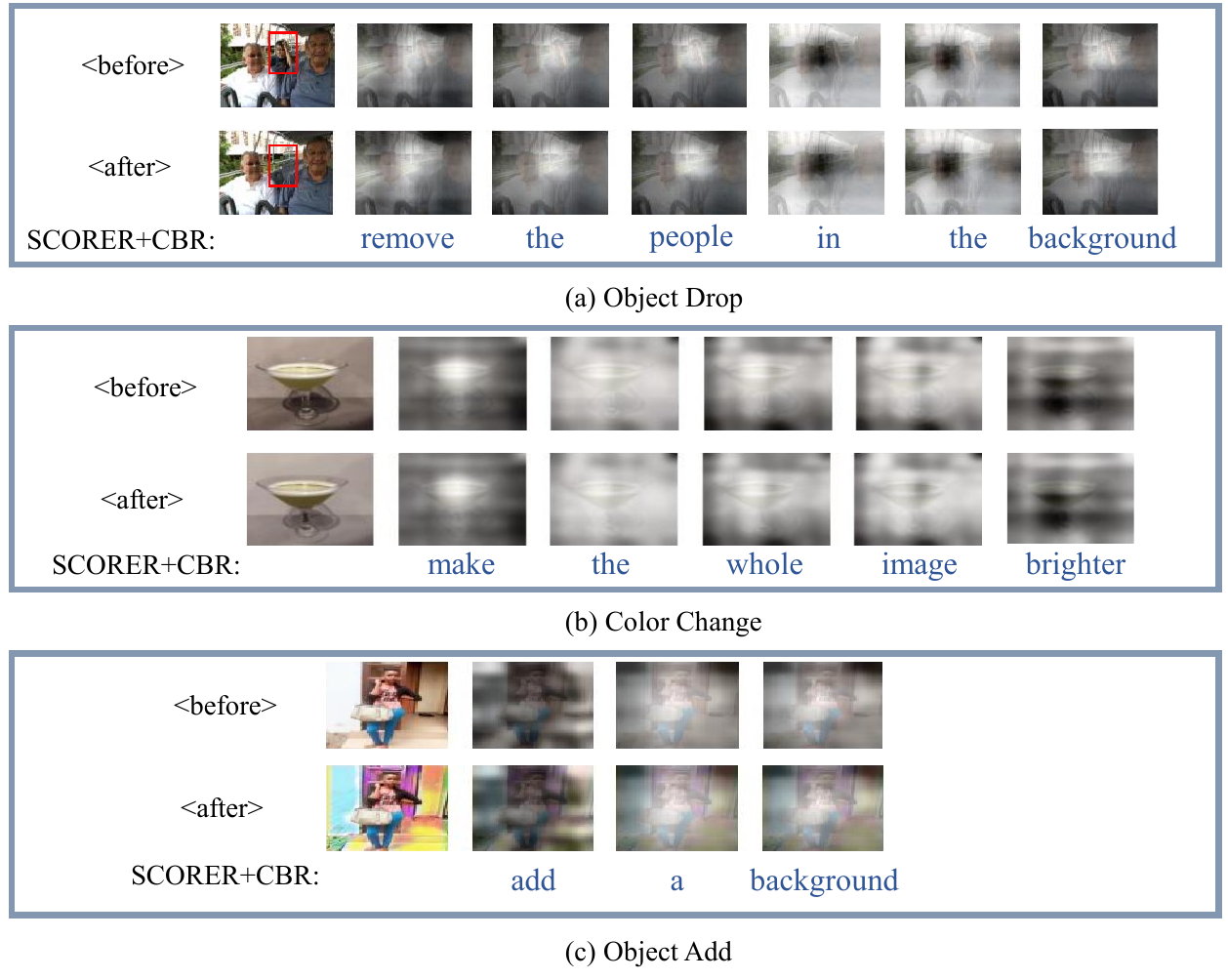} 
\caption{Three cases about ``Drop'', ``Color Change'', and ``Add'' from Image Editing Request, where the generated captions along with the attention weight  at each word are visualized.  }
\label{word_dc}
\end{figure*}

\begin{figure*}[htbp]
\centering
\includegraphics[width=0.9\textwidth]{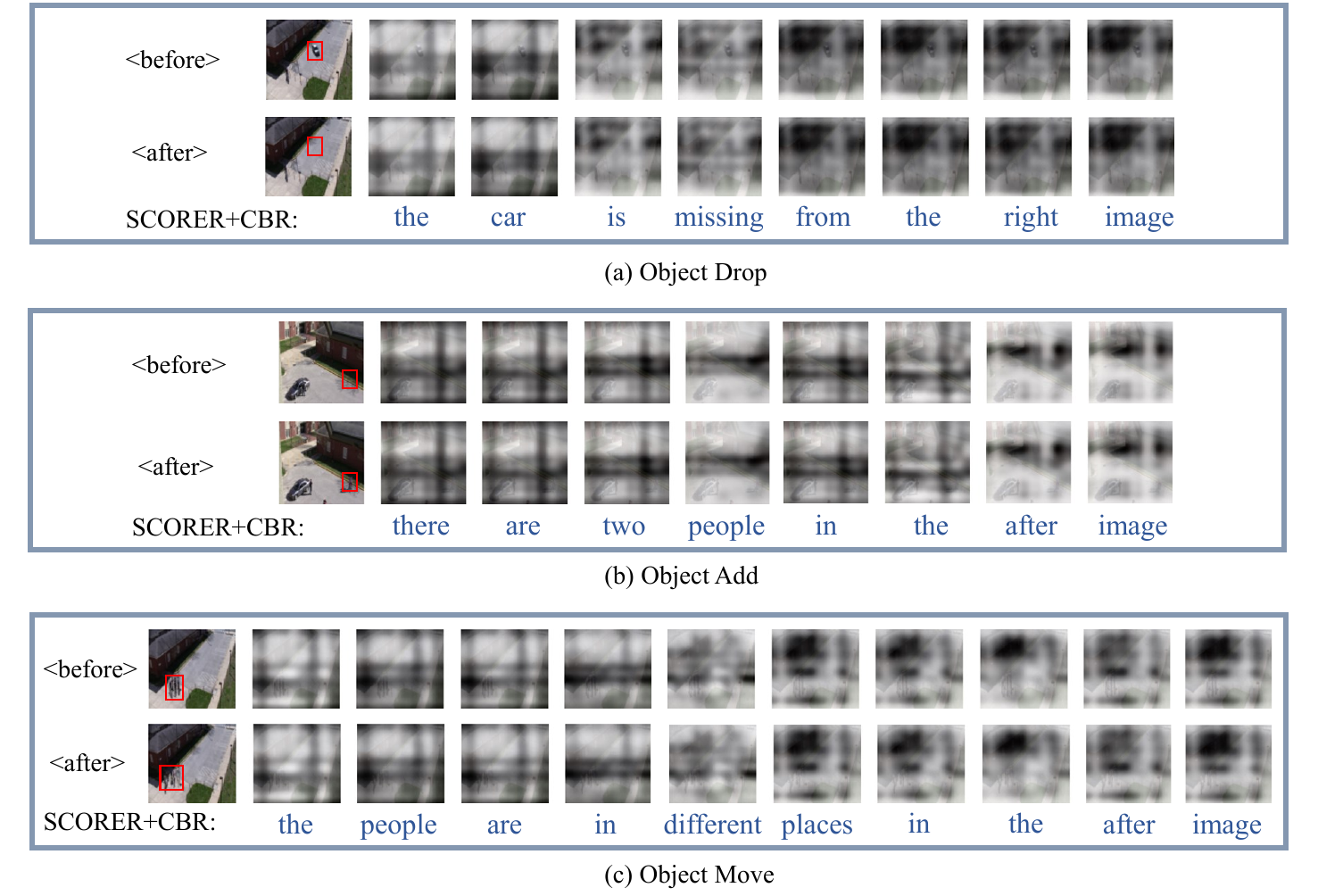} 
\caption{Three cases about ``Drop'', ``Add'', and ``Move'' from Spot-the-Diff, where the generated captions along with the attention weight at each word are visualized.  }
\label{word_spot}
\end{figure*}

\end{document}